\documentclass{article}

\newcommand{\ourmethod}{TextPro-SLM}
\newcommand{\ourencoder}{WhisperPro}

\PassOptionsToPackage{numbers,sort&compress}{natbib}

 \usepackage[main, final]{neurips_2026}

\usepackage[utf8]{inputenc} 
\usepackage[T1]{fontenc}    
\usepackage{hyperref}       
\usepackage{url}            
\usepackage{booktabs}       
\usepackage{multirow}       
\usepackage{amsfonts}       
\usepackage{nicefrac}       
\usepackage{microtype}      
\usepackage[table]{xcolor} 
\usepackage{enumitem}
\usepackage{pgf}
\usepackage{xparse}
\usepackage{wrapfig}

\definecolor{lightblue}{RGB}{230,243,255}

\newcommand{\applycellcolor}[1]{%
  \begingroup\edef\gap@tmp{\endgroup\noexpand\cellcolor{#1}}\gap@tmp}

\newcommand{\GapTone}{15} 
\colorlet{GapGreen}{green!\GapTone!white}
\colorlet{GapRed}{red!\GapTone!white}

\NewDocumentCommand{\GapCell}{m m m m}{%
  \pgfmathparse{max(0,min(100,round(100*(#1-#2)/(#3-#2))))}%
  \applycellcolor{GapGreen!\pgfmathresult!GapRed}#4%
}

\NewDocumentCommand{\GapSC}{m O{#1}}   {\GapCell{#1}{20}{0.0}{#2}}
\NewDocumentCommand{\GapMMSU}{m O{#1}} {\GapCell{#1}{20}{0.0}{#2}}
\NewDocumentCommand{\GapOBQA}{m O{#1}} {\GapCell{#1}{20}{0.0}{#2}}
\NewDocumentCommand{\GapARCE}{m O{#1}} {\GapCell{#1}{20}{0.0}{#2}}
\NewDocumentCommand{\GapARCC}{m O{#1}} {\GapCell{#1}{20}{0.0}{#2}}
\NewDocumentCommand{\GapPIQA}{m O{#1}} {\GapCell{#1}{20}{0.0}{#2}}
\NewDocumentCommand{\GapAVG}{m O{#1}}  {\GapCell{#1}{20}{0.0}{#2}}

\newtheorem{remark}{Remark}
\usepackage{graphicx}

\title{Minimizing Modality Gap from the Input Side: Your Speech LLM Can Be a Prosody-Aware Text LLM}

\author{
 \textbf{Wenqian Cui\textsuperscript{1}}\thanks{Contact: wenqian.cui@link.cuhk.edu.hk}, 
 \textbf{Xiao-Hui Li\textsuperscript{2}}\thanks{Corresponding author.}, 
 \textbf{Daxin Tan\textsuperscript{2}}, 
 \textbf{Qiyong Zheng\textsuperscript{1}}, 
 \textbf{Irwin King\textsuperscript{1}}\footnotemark[2]
\\
 \textsuperscript{1}The Chinese University of Hong Kong,
 \textsuperscript{2}Huawei Technologies
}

\begin{document}

\maketitle

\begin{abstract}
  Speech large language models (SLMs) are typically built from text large language model (TLM) checkpoints, yet they still suffer from a substantial modality gap. Prior work has mainly attempted to reduce this gap from the output side by making speech generation more text-like, but the gap remains. We argue that the key remaining bottleneck lies on the input side. We propose \ourmethod{}, an SLM that makes spoken input more closely resemble that of a prosody-aware text LLM. \ourmethod{} combines \ourencoder{}, a unified speech encoder that produces synchronized text tokens and prosody embeddings, with an LLM backbone trained to preserve the semantic capabilities of the original TLM while learning paralinguistic understanding. Experiments show that \ourmethod{} achieves the lowest modality gap among leading SLMs at both 3B and 7B scales, while also delivering strong overall performance on paralinguistic understanding tasks. These gains are achieved with only roughly 1{,}000 hours of LLM training audio, suggesting that reducing the modality gap from the input side is both effective and data-efficient.\footnote{Code is available at \url{https://github.com/dreamtheater123/TextPro-SLM}.}
\end{abstract}

\section{Introduction}
\label{sec:intro}
Speech Large Language Models (Speech LLMs, SLMs) have been extensively studied in recent years to enable seamless speech-based interaction \cite{SLMsurvey,llama-omni,moshi,turnguide}. Most SLMs are built by continually training from Text Large Language Model (TLM) checkpoints \cite{gpt4,gpt5,llama4,gemini,gemini1_5,glm2024chatglm}, with the goal of extending the strong reasoning and language understanding capabilities of TLMs to spoken input and output. However, despite this shared backbone, a major challenge remains: the \textbf{modality gap}---speech-based question answering (QA) performance often remains substantially worse than the text-based QA performance of the underlying TLM, limiting the practical usability of SLMs. For example, prior studies \cite{salad} report that GLM-4-Voice \cite{glm-4-voice} suffers up to a 20\% performance drop on several QA benchmarks.


Most existing studies attempt to bridge the modality gap from the \textbf{output side}. Early SLMs use the LLM backbone to generate only speech tokens, enabling fully end-to-end spoken interaction \cite{OnGenerativeRawAudio,twist,spiritlm}. Later work improves performance by having the LLM first generate intermediate text tokens before emitting speech tokens \cite{speechgpt,mini-omni,glm-4-voice,vita-audio}. More recent approaches further decouple text generation from speech synthesis through a ``thinker-talker'' architecture, in which the LLM backbone generates only text tokens while a separate talker module produces speech conditioned on the thinker's hidden states \cite{llama-omni2,minmo,vocalnet,qwen3-omni,qwen25-omni}. This line of progress reveals a clear pattern: the closer the generation process is to the interaction paradigm of a TLM, the better the resulting SLM tends to perform. Nevertheless, even these advances do not eliminate the modality gap.

\begin{figure}[t]
  \centering
  \includegraphics[width=1.0\textwidth]{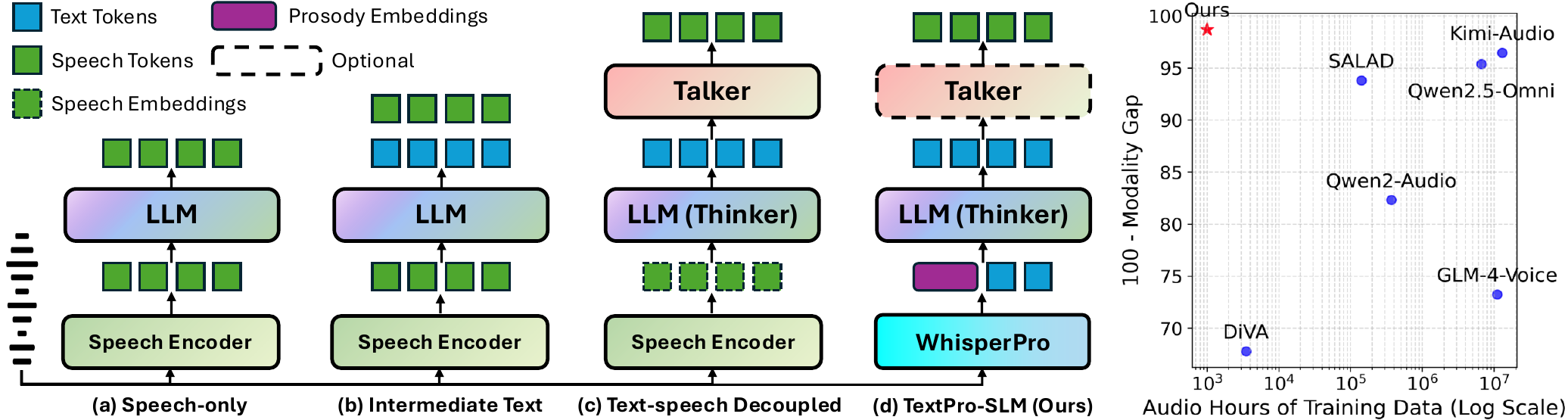}
  \caption{Left: Comparison of the architectural designs used by prior SLMs and our approach. Right: Comparison of modality gap versus the amount of audio training data across different approaches.}
  \label{fig:intro_figure}
\end{figure}


This observation suggests that the remaining bottleneck may lie primarily on the \textbf{input side}. Under modern thinker-talker designs, the output side already resembles that of a TLM. The more fundamental difficulty is that, unlike a TLM, an SLM must infer both linguistic content and paralinguistic cues directly from speech representations, which are often not presented to the LLM in a form it can fully exploit \cite{gap_understanding_direction_magnitude,soundwave}. 
Motivated by this perspective, we propose to minimize the modality gap from the input side by making a Speech LLM behave as closely as possible to a \textbf{prosody-aware text LLM}. The central idea is simple: if the output side of modern SLMs is already converging toward text-style generation, then the input side should also provide the backbone with a representation that is both TLM-compatible and prosody-preserving. Rather than treating speech purely as an acoustic signal that must be compressed into a single latent stream, we explicitly separate and preserve two complementary aspects of spoken input: \emph{what is said} and \emph{how it is said}.

To achieve this, we propose \textbf{\ourmethod{}}, which consists of two key components. \textbf{1) Speech encoder.} We introduce \textbf{\ourencoder{}}, a unified \textbf{Whisper}-based speech encoder that produces two outputs: text tokens that capture utterance content and \textbf{Pro}sody embeddings that encode paralinguistic information such as speaking style, emotion, and speaker timbre. \ourencoder{} is trained to support both accurate transcription and prosody-rich representation learning, so the downstream LLM receives a richer and more TLM-compatible view of the input speech. \textbf{2) LLM backbone.} We train the LLM backbone with two coupled objectives: preserving the semantic capabilities of the original TLM and learning to understand paralinguistics. Specifically, we retain the semantic intelligence of the underlying TLM through knowledge distillation, while jointly optimizing the model on paralinguistic-understanding tasks. In this way, both linguistic meaning and paralinguistic information can propagate through the model, yielding an end-to-end SLM that remains semantically strong while also interpreting how something is spoken. Figure \ref{fig:intro_figure} compares the architectural design of \ourmethod{} with prior solutions.

Experimental results show that our method not only achieves the lowest modality gap among the compared approaches but also attains competitive performance on paralinguistic-understanding tasks, while remaining extremely \textbf{data efficient}, requiring only roughly 1,000 hours of audio data for LLM training. We attribute this efficiency to the fact that our design stays closer to the original operating regime of the underlying TLM than prior SLM architectures, making adaptation from text to speech more direct and less data-intensive. Overall, our results suggest that an effective route toward stronger Speech LLMs is to turn them into prosody-aware text LLMs.


\section{Related Work}
\label{sec:relatedwork}
\subsection{Speech Large Language Models}
SLMs aim to enable seamless spoken dialogue interactions \cite{SLMsurvey}. A typical SLM consists of three main components: 1) a speech encoder that converts the input waveform into speech representations, typically either continuous speech embeddings \cite{qwen3-omni,qwen25-omni} or discrete speech tokens \cite{wav2vec20,w2v-bert,hubert}; 2) an LLM backbone that processes these representations and generates outputs, which may be text tokens \cite{qwen25-omni}, speech tokens \cite{twist}, or both \cite{glm-4-voice}; and 3) a speech decoder that synthesizes the model output into spoken responses. Early SLMs are trained purely on speech data and typically represent speech as fully discrete tokens \cite{OnGenerativeRawAudio}. Later work shows that leveraging a pre-trained TLM as the backbone leads to substantially stronger performance on speech tasks \cite{twist}. Since then, building SLMs from TLM checkpoints has become standard practice, with much of the research focusing on explicit speech-text alignment leveraging Automatic Speech Recognition (ASR), Text-to-Speech Synthesis (TTS), or interleaved speech-text training data \cite{spiritlm,glm-4-voice,kimi-audio}.

Recent SLMs can be broadly grouped into two architectural paradigms according to the form of speech input provided to the LLM. \textbf{1) Discrete models:} the speech encoder produces discrete speech tokens, and the LLM consumes these tokens directly to generate text tokens, speech tokens, or both \cite{glm-4-voice,vita-audio}. \textbf{2) Continuous models:} the speech encoder produces continuous embeddings, which are then mapped into the LLM embedding space through a projector module. Models in this category often adopt a thinker-talker architecture, where the thinker generates text tokens and a separate talker produces speech conditioned on the thinker's hidden states \cite{qwen3-omni,qwen25-omni}. Our approach differs from both paradigms by providing the LLM with text tokens together with prosody embeddings, rather than asking it to infer both semantic and prosodic information from a single speech representation. This design better preserves the operating regime of the underlying TLM and helps reduce the text-speech modality gap effectively.

\subsection{Text-Speech Modality Gap}
As discussed in Section \ref{sec:intro}, recent SLM architectures have narrowed the modality gap mainly by improving the output side. Here, we review work that explicitly studies this gap and work that attempts to reduce it, then clarify how our approach differs.

Some studies focus on understanding the gap itself. Hsu \textit{et al.} \cite{anatomy_gap_understanding_long_parse} attribute it to the longer speech sequence and the sparsity of semantic information in speech. Xiang et al. \cite{gap_understanding_direction_magnitude} analyze the gap at the representation level and show that speech and text representations are similar in direction but differ in magnitude. Other studies aim to reduce the gap, mostly through improved training. SALAD \cite{salad} and Wang \textit{et al.} \cite{cross_modal_distillation} use Kullback-Leibler (KL) divergence to align the student's output on speech input with the teacher's output on text input. X-OPD \cite{x-opd} and CORD \cite{cord} extend this line with on-policy distillation, while Yang \textit{et al.} \cite{layerwise_distill} perform finer-grained layer-wise distillation. A separate line of work targets catastrophic forgetting during continual training on speech data: DeepOmni \cite{deepomni} uses a Mixture-of-Experts architecture \cite{MoE_survey} that separates text and speech experts, and Hsiao \textit{et al.} \cite{gap_continual_learning} identify experience replay as the most effective continual-learning strategy \cite{multimodal_continual_learning_survey}. However, these methods mainly improve optimization and training stability rather than the interaction paradigm itself, which we view as a more fundamental source of the modality gap.

\begin{figure}[t]
  \centering
  \includegraphics[width=1.0\textwidth]{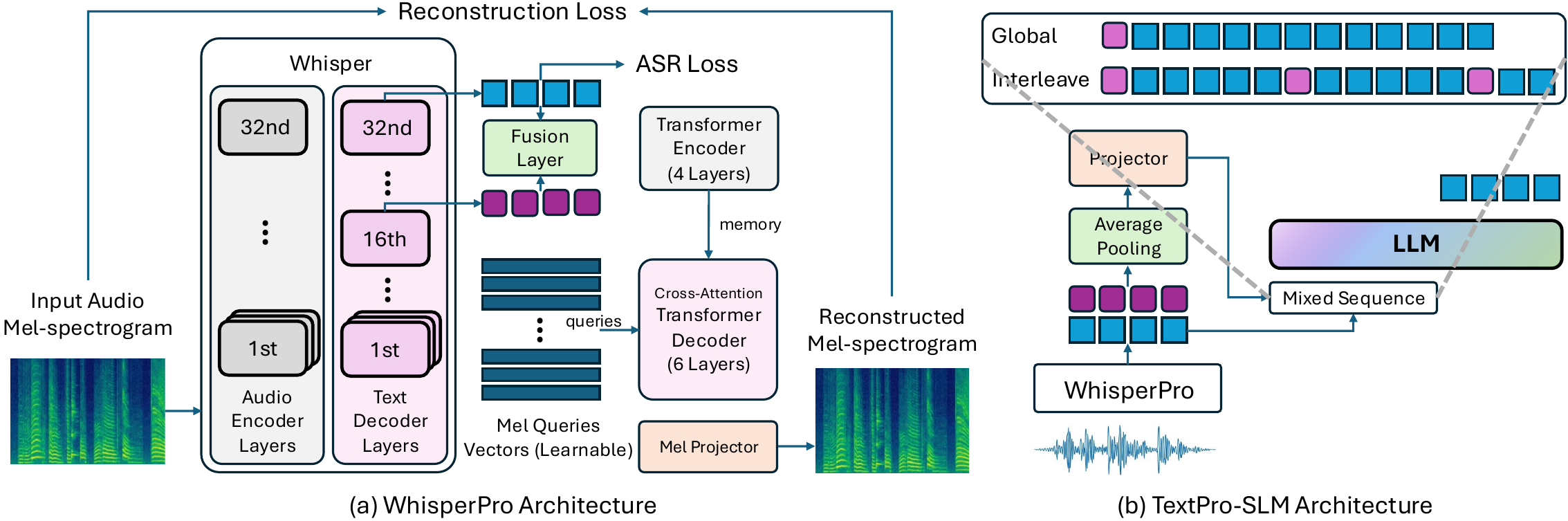}
  \caption{Model architecture of \ourencoder{} and \ourmethod{}.}
  \label{fig:methodology}
\end{figure}

\section{Methodology}
\label{sec:method}
Our approach to minimizing the modality gap has two core components. Section \ref{sec:speechencoder} presents \ourencoder{}, our speech encoder, and Section \ref{sec:llmbackbone} introduces the prosody-aware LLM backbone.

\subsection{\ourencoder{}: An ASR Speech Encoder that Encodes Everything}
\label{sec:speechencoder}
We design a speech encoder that outputs both accurate transcripts and prosody-rich embeddings. Broadly, there are three ways to build such a model: starting from an ASR model and adding paralinguistic modeling, starting from a prosody-rich model such as a neural audio codec and adding transcription ability, or training a new model from scratch for both goals. We adopt the first strategy for two reasons. First, prior work suggests that SLMs rely more heavily on semantic information than on fine-grained acoustic details \cite{audiolm,hussein2025hasrd}. Second, strong ASR models are typically trained on much larger datasets than codec models. For instance, Whisper-large-v3 \cite{whisper} is trained on 680,000 hours of audio, whereas a competitive codec model such as WavTokenizer \cite{wavtokenizer} only requires as few as 585 hours. Starting from a strong ASR backbone therefore fits the downstream LLM well and requires significantly less training data. In practice, we choose Whisper-large-v3 \cite{whisper} because of its strong transcription performance and robustness across diverse acoustic conditions.

The main challenge of this task is that ASR models are trained to prioritize lexical content, not prosody. To inject paralinguistic information into the representation space, we borrow the core idea behind codec models: preserve acoustic details through reconstruction. Concretely, we use Whisper as the encoder to produce both text tokens and token-aligned prosody embeddings, and we attach an additional decoder that reconstructs the input speech from these two outputs. This design encourages the encoder states to retain not only what was said, but also how it was said. Figure \ref{fig:methodology} illustrates the overall architecture of \ourencoder{}.

\textbf{Architecture.}
The architecture of \ourencoder{} consists of two modules: Whisper-large-v3 and a mel-reconstructor. Let the input utterance be a waveform \(\mathbf{x}\in\mathbb{R}^T\). We first follow Whisper's standard preprocessing pipeline to obtain a log-mel spectrogram \(\mathbf{M}\in\mathbb{R}^{F\times L}\).
Whisper-large-v3 then maps \(\mathbf{M}\) to both the transcript and the prosody embeddings. Specifically, Whisper autoregressively produces a text token sequence
\begin{equation}
\mathbf{y}=\mathrm{Whisper}(\mathbf{M})=(y_1,\dots,y_N).
\label{eq:whisper-transcript}
\end{equation}
Whisper-large-v3 consists of 32 Transformer encoder layers and 32 Transformer decoder layers. We take the hidden states of each text token from the 16-th Whisper decoder layer as token-aligned prosody embeddings, denoted as
\begin{equation}
\mathbf{P}=\mathrm{WhisperDec}^{(16)}(\mathbf{M},\mathbf{y})=(p_1,\dots,p_N)\in\mathbb{R}^{N\times d}.
\label{eq:whisper-prosody}
\end{equation}
Because these states are aligned with the generated text tokens, \ourencoder{} represents speech as two synchronized streams, \(\{(y_i,\mathbf{p}_i)\}_{i=1}^N\), capturing \emph{what} was said and \emph{how} it was said. To force the model to preserve prosodic information, we attach a mel-reconstructor that predicts the original mel spectrogram from the transcript-prosody pairs. At the overview level, we denote this module as
\begin{equation}
\hat{\mathbf{M}}=\mathrm{MelReconstructor}(\mathbf{y},\mathbf{P}).
\label{eq:mel-recon-overview}
\end{equation}
Concretely, the text tokens and prosody embeddings are mapped into a shared space, fused token-wise, contextualized by a Transformer encoder, and decoded into frame-level acoustic features by a Transformer decoder with learnable frame queries. The detailed architecture of the mel-reconstructor is presented in Appendix \ref{apx:mel-reconstructor_details}. This reconstruction objective encourages the decoder states to retain information not only for transcription, but also for recovering the acoustic realization of the utterance.

During training, \ourencoder{} is optimized with a multi-task objective consisting of an ASR loss and a Mel reconstruction loss. Let \(\mathbf{y}^*=(y^*_1,\dots,y^*_N)\) denote the ground-truth transcription. The ASR objective is the average token-level cross-entropy:
\begin{equation}
\mathcal{L}_{\mathrm{ASR}}
=
-\frac{1}{N}\sum_{i=1}^{N}\log p(y_i^*\mid y_{<i}^*, \mathbf{M}),
\label{eq:loss-asr-main}
\end{equation}
and the mel-reconstructor is trained with a mean squared error loss between the reconstructed and original Mel spectrograms:
\begin{equation}
\mathcal{L}_{\mathrm{mel}}
=
\frac{1}{FL}
\left\|
\hat{\mathbf{M}}-\mathbf{M}
\right\|_2^2.
\label{eq:loss-mel-main}
\end{equation}
The final training objective is
\begin{equation}
\mathcal{L}
=
\mathcal{L}_{\mathrm{ASR}}
+
\lambda\,\mathcal{L}_{\mathrm{mel}},
\label{eq:loss-total-main}
\end{equation}
where \(\lambda\) is a hyperparameter and is empirically set to $1.0$.

\textbf{Training.}
We use the 960-hour LibriSpeech training split \cite{librispeech} as our primary training corpus. Because the original LibriSpeech transcripts are written in uppercase and do not include punctuation, we pair each audio utterance with its punctuated transcription from LibriSpeech-PC \cite{librispeech-pc}. In preliminary experiments, we observe that Whisper becomes less stable when transcribing utterances that contain both questions and answer choices. To improve ASR robustness in QA settings, we further augment the training data with speech versions of CommonsenseQA \cite{commonsenseqa}, synthesized using Kokoro TTS with the af-heart voice \footnote{\url{https://github.com/hexgrad/kokoro-82M}}. We train \ourencoder{} for 2 epochs with a global batch size of 16, a learning rate of $1\times10^{-5}$, and a linear decay scheduler with 500 warmup steps.

\subsection{Prosody-Aware LLM}
\label{sec:llmbackbone}
Our LLM backbone is designed to simultaneously minimize the semantic modality gap and preserve strong paralinguistic understanding. We build on instruction-tuned TLMs, specifically Qwen2.5-3B-Instruct and Qwen2.5-7B-Instruct at different model scales.

\begin{wraptable}{r}{0.46\textwidth}
  \vspace{-1.5\baselineskip}
  \centering
  \caption{\label{tab:dataset_summary}
    Summary of the datasets used in \ourencoder{} and LLM training, where Recon, KD, SER, and AGE represent Reconstruction, Knowledge Distillation, Speech Emotion Recognition, and Age Detection, respectively.
  }
  \setlength{\tabcolsep}{3.5pt}
  \renewcommand{\arraystretch}{0.95}
  \resizebox{\linewidth}{!}{
    \begin{tabular}{lcc}
      \toprule
      \textbf{Datasets} & \textbf{Objective} & \textbf{Duration (h)} \\
      \midrule
      \rowcolor{lightblue} \multicolumn{3}{c}{\textbf{\ourencoder{}}} \\
      CommonsenseQA \cite{commonsenseqa} & ASR + Recon & 40 \\
      LibriSpeech \cite{librispeech} & ASR + Recon & 960 \\
      \textbf{Total}  & - & 1,000 \\
      \midrule
      \rowcolor{lightblue} \multicolumn{3}{c}{\textbf{LLM}} \\
      CommonsenseQA \cite{commonsenseqa} & KD & 40 \\
      UltraChat (part) \cite{ultrachat} & KD & 255 \\
      ParaSpeechCaps \cite{paraspeechcaps} & Captioning & 342 \\
      IEMOCAP \cite{iemocap} & SER & 12 \\
      CREMA-D \cite{crema-d} & SER & 5 \\
      SAVEE \cite{savee} & SER & 0.5 \\
      TESS \cite{tess} & SER & 1.6 \\
      ESD (Eng) \cite{esd} & SER & 14.5 \\
      Common Voice (part) \cite{commonvoice} & AGE & 319 \\
      \textbf{Total} & - & 989.6 \\
      \bottomrule
    \end{tabular}
  }
  \vspace{-1.0\baselineskip}
\end{wraptable}
\textbf{Input Format.}
Our key design principle is to convert speech into a representation that better matches the native input format of a text LLM, while keeping semantic information dominant. Given the transcript and prosody embeddings produced by \ourencoder{}, we feed the transcript directly into the LLM as the primary input and inject a compressed prosody representation alongside it. We consider two input organization strategies:
\textbf{1) Global prepending (default):} We summarize prosody into a single utterance-level vector and prepend it to the textual input. As we show later, this compact representation is already sufficient for strong paralinguistic understanding.
\textbf{2) Interleaving:} To account for cases where prosodic changes within an utterance may matter, we additionally provide an interleaved formulation that inserts compressed prosody embeddings among text tokens at a fixed ratio.
These designs preserve the original capabilities of the LLM while still providing a compact representation of how the utterance was spoken, consistent with prior findings that semantic information plays a more central role than raw acoustic detail in SLM modeling \cite{audiolm,hussein2025hasrd}.

Let the \ourencoder{} produce a transcription token sequence \(\mathbf{y}^{\mathrm{enc}}=(y^{\mathrm{enc}}_1,\dots,y^{\mathrm{enc}}_N)\) and the aligned prosody embeddings
\(\mathbf{P}=(p_1,\dots,p_N).\) We then tokenize the transcribed text again using the LLM tokenizer, yielding
\(\mathbf{y}^{\mathrm{llm}}=(y^{\mathrm{llm}}_1,\dots,y^{\mathrm{llm}}_T)\), where in general \(T\neq N\) because the tokenization schemes of \ourencoder{} and the LLM need not coincide.
We instantiate two ways of injecting prosody into the LLM, corresponding to the global prepending and interleaving designs introduced above.
For \textbf{global prepending}, we first aggregate all prosody embeddings from the user question into a single utterance-level vector by average pooling:
\begin{equation}
\bar{\mathbf{p}}=\frac{1}{N}\sum_{i=1}^{N}p_i \in \mathbb{R}^{d}.
\label{eq:prosody-pooling}
\end{equation}
We then project the averaged prosody embedding into the LLM embedding space using a two-layer Multi-Layer Perceptron (MLP), and place this projected prosody embedding at the front of the input sequence. Therefore, the final input to the LLM is
\begin{equation}
  \label{eq:llm_input_seq_global}
  \mathbf{H}^{(0)}_{\mathrm{global}}=\bigl[\tilde{\mathbf{p}};\, y^{\mathrm{llm}}_1;\, \dots;\, y^{\mathrm{llm}}_T\bigr],
\end{equation}
where $\tilde{\mathbf{p}}$ denotes the projected global prosody embedding. Thus, the backbone receives a sequence that is almost identical to a standard text-only prompt, except for one prepended prosody token summarizing how the question was spoken.
For \textbf{interleaving}, we preserve finer-grained prosodic variation by first determining how many prosody embeddings will be inserted into the LLM input, and then compressing the original prosody sequence into that many averaged embeddings. Given a fixed insertion ratio \(r\), we calculate \(M=\lceil T/r \rceil\) as the number of interleaved groups needed for the LLM token sequence, and then evenly partition the original prosody sequence \(\mathbf{P}\) of length \(N\) into \(M\) consecutive groups. For each group, we average the prosody embeddings within that group and apply the same projector to obtain one interleaved prosody embedding \(\tilde{p}_j\), resulting in \(\tilde{\mathbf{P}}=(\tilde{p}_1,\dots,\tilde{p}_M)\). Finally, the projected prosody embeddings and text tokens are interleaved, giving the final input sequence represented as
\begin{equation}
\mathbf{H}^{(0)}_{\mathrm{interleave}}=
\bigl[\tilde{p}_1;\, \mathbf{y}^{(1)};\, \tilde{p}_2;\, \mathbf{y}^{(2)};\, \dots;\, \tilde{p}_M;\, \mathbf{y}^{(M)}\bigr],
\label{eq:llm_input_seq_interleave}
\end{equation}
where \(\mathbf{y}^{(j)}\) denotes the \(j\)-th LLM text-token group, containing up to \(r\) tokens.
\begin{remark}
We omit the training of the speech synthesis module in this work due to the following two reasons. 1) Dominant approaches in this field measure the modality gap using textual outputs. 2) We see speech synthesis based on textual outputs as a largely decoupled problem whose primary goal is to improve speech naturalness and fidelity, rather than to reduce the modality gap between speech understanding and text understanding.
\end{remark}

\begin{table}[t]
\centering
\caption{ASR and paralinguistic probing performance of \ourencoder{} compared to Whisper-large-v3. Probing results are reported for 1-layer and 2-layer MLP probes.}
\label{tab:encoder_results_merged}
\small
\begingroup
\setlength{\tabcolsep}{4pt}
\begin{tabular}{llcccccccc}
\toprule
\multirow{2}{*}{\textbf{Model}} & \multirow{2}{*}{\textbf{Probe}} & \multicolumn{2}{c}{\textbf{WER} $\downarrow$} & \multicolumn{4}{c}{\textbf{Probing} $\uparrow$} & \multirow{2}{*}{\textbf{Avg.}} \\
& & \textbf{LibriSpeech} & \textbf{GigaSpeech} & \textbf{SER} & \textbf{SI} & \textbf{F0} & \textbf{Energy} & \\
\midrule
\multirow{2}{*}{Whisper-large-v3}
& 1-layer & \multirow{2}{*}{4.43\%} & \multirow{2}{*}{13.66\%} & 33.84 & 38.83 & 27.45 & 14.64 & 28.69 \\
& 2-layer &  &  & 41.02 & 60.67 & 34.09 & 18.35 & 38.53 \\
\midrule
\multirow{2}{*}{\ourencoder{}}
& 1-layer & \multirow{2}{*}{\textbf{1.96\%}} & \multirow{2}{*}{\textbf{12.89\%}} & 53.52 & 79.31 & 46.81 & 30.78 & 52.61 \\
& 2-layer &  &  & 69.63 & 87.84 & 50.34 & 36.56 & 61.09 \\
\bottomrule
\end{tabular}
\endgroup
\end{table}

\textbf{Training Data and Pipeline.} We utilize two sets of training data for different purposes.
\begin{enumerate}[left=0pt]
  \item \textbf{Semantic Intelligence Preservation:} Following prior work \cite{salad,cross_modal_distillation}, we construct training data through knowledge distillation from teacher models. Specifically, we use UltraChat\footnote{We use only 1/10 of UltraChat, i.e., \texttt{train\_2.jsonl}, from \url{https://huggingface.co/datasets/openbmb/UltraChat}.} \cite{ultrachat} to distill general instruction-following ability and CommonsenseQA \cite{commonsenseqa} to distill QA ability. We use the original TLMs as teacher models, and the distillation pipeline is as follows. First, we use Kokoro TTS to convert the text-only datasets into speech. Next, we pass the synthesized speech through \ourencoder{} to obtain transcriptions and prosody embeddings. We then use the transcriptions to query the teacher models and collect their responses. Finally, we feed the transcription-prosody pairs into the LLM backbone as inputs and use the teacher responses as supervision targets.

  \item \textbf{Paralinguistic Understanding Capabilities:} To develop paralinguistic understanding, we focus on four representative speech attributes: speaker emotion, gender, accent, and age. We use two types of datasets for paralinguistic training. \textbf{1) Speech Captioning:} Speech captioning datasets pair speech utterances with natural-language descriptions that often capture paralinguistic cues. Training on such data helps the model learn to associate speech signals with the paralinguistic information they convey. We use the ParaSpeechCaps dataset \cite{paraspeechcaps} for this purpose since it contains the gender and accent attributes. 
  \textbf{2) Feature Detection:} Feature detection datasets provide speech utterances together with ground-truth annotations for a target paralinguistic attribute. Because the captions in ParaSpeechCaps do not include emotion or age labels, we use dedicated feature detection datasets to train the model on these two attributes. For emotion, we use IEMOCAP \cite{iemocap}, CREMA-D \cite{crema-d}, SAVEE \cite{savee}, TESS \cite{tess}, and the English portion of ESD \cite{esd}. For age, we use the first 1/4 of the Common Voice data \cite{commonvoice} with speaker-age labels.

  \begin{remark}
  We acknowledge that, although the features described above are representative, they do not cover the full paralinguistic space. Nevertheless, we use them as a proof of concept to show that our model can demonstrate strong paralinguistic understanding capabilities.
  \end{remark}
\end{enumerate}

We adopt a two-stage training pipeline. In the first stage, only the prosody projector is trained while the LLM backbone remains frozen. In the second stage, we jointly train the projector and the LLM. 
Each stage trains for 2 epochs with a global batch size of 1,024 and a linear decay scheduler with no warmups. We use a learning rate of $1\times10^{-4}$ in Stage~1 and $1\times10^{-5}$ in Stage~2. Both stages are trained with cross-entropy loss.
Table \ref{tab:dataset_summary} summarizes the datasets used to train \ourencoder{} and the LLM, along with their training objectives and audio durations.

\section{Experiments}
\label{sec:experiments}
We conduct experiments to demonstrate that our proposed approach minimizes the modality gap while demonstrating strong paralinguistic understanding abilities. We provide both \ourencoder{} and LLM evaluation results.

\subsection{\ourencoder{} Evaluations}
Before evaluating the full system, we first assess \ourencoder{} in isolation. Our goal is to verify that it preserves Whisper's strong semantic modeling ability while learning richer prosodic representations. To this end, we compare \ourencoder{} directly with the original Whisper-large-v3 along two dimensions: ASR quality and paralinguistic information encoded in the prosody embeddings.

To evaluate semantic modeling, we measure ASR performance on the test sets of LibriSpeech \cite{librispeech} and GigaSpeech \cite{gigaspeech} using word error rate (WER). As shown in Table~\ref{tab:encoder_results_merged}, \ourencoder{} achieves slightly lower WER than Whisper-large-v3 on both benchmarks. This result indicates that our modifications do not compromise the strong transcription ability of the original Whisper model.

To evaluate prosodic modeling, we conduct probing experiments on four paralinguistic attributes: speech emotion recognition (SER), speaker identification (SI), fundamental frequency (F0), and energy. We use RAVDESS \cite{ravdess} for SER, the VoxCeleb1 test set \cite{voxceleb1} for SI, and the LibriTTS test set \cite{libritts} for F0 and energy. Following standard probing practice, we train both 1-layer and 2-layer MLP classifiers on top of the extracted prosody embeddings. Because F0 and energy are continuous-valued signals rather than categorical labels, we discretize each feature into 10 quantile-based bins. More experimental details of the probing experiments are presented in Appendix \ref{apx:probing}. Table~\ref{tab:encoder_results_merged} shows that the prosody embeddings produced by \ourencoder{} consistently encode substantially richer information than the original Whisper representations across all four attributes. Together, these findings suggest that our reconstruction objective significantly enhances paralinguistic representation learning while preserving semantic fidelity.

\begin{table}[t]
    \caption{Various SLM performance $\uparrow$ (\%) and modality gap $\downarrow$ on different benchmarks. Gap cells are color-coded from best (green) to worst (red) for each task, where lower is better. We save the word Instruct for Kimi-Audio, Qwen2.5-3B and 7B to save spaces. 
    }
    \label{tab:modality_gap}
    \begin{center}
    \resizebox{\textwidth}{!}{
        \begin{tabular}{@{}llllllllllllllll@{}}
\toprule
\multirow{2}{*}{} & \multicolumn{2}{c}{StoryCloze} & \multicolumn{2}{c}{MMSU} & \multicolumn{2}{c}{OBQA} & \multicolumn{2}{c}{ARC-Easy} & \multicolumn{2}{c}{ARC-C} & \multicolumn{2}{c}{PIQA} & \multicolumn{2}{c}{Avg.} \\
\cmidrule(l){2-13}\cmidrule(l){14-15}
& \multicolumn{1}{c}{Acc.} & \multicolumn{1}{c}{Gap} 
& \multicolumn{1}{c}{Acc.} & \multicolumn{1}{c}{Gap} 
& \multicolumn{1}{c}{Acc.} & \multicolumn{1}{c}{Gap} 
& \multicolumn{1}{c}{Acc.} & \multicolumn{1}{c}{Gap} 
& \multicolumn{1}{c}{Acc.} & \multicolumn{1}{c}{Gap} 
& \multicolumn{1}{c}{Acc.} & \multicolumn{1}{c}{Gap} 
& \multicolumn{1}{c}{Acc.} & \multicolumn{1}{c}{Gap} \\
\midrule
Random & 50.0 & \multicolumn{1}{c}{-} & 25.0 & \multicolumn{1}{c}{-} & 25.0 & \multicolumn{1}{c}{-} & 25.0 & \multicolumn{1}{c}{-} & 25.0 & \multicolumn{1}{c}{-} & 50.0 & \multicolumn{1}{c}{-} & 33.3 & - \\
\midrule
\multicolumn{15}{c}{\textit{Cascaded \textbf{Toplines}: ASR (Whisper-v3-Large) + LLM}} \\
\midrule
ASR + Qwen2.5-3B   & 88.3 & \GapSC{4.3} & 58.9 & \GapMMSU{2.7}[\textbf{2.7}] & 72.7 & \GapOBQA{4.0} & 90.5 & \GapARCE{3.7} & 80.1 & \GapARCC{3.7} & 52.7 & \GapPIQA{-0.6} & 73.9 & \GapAVG{3.0} \\
ASR + Qwen2.5-7B   & \textbf{89.1} & \GapSC{0.1}[\textbf{0.1}] & \textbf{67.9} & \GapMMSU{2.9} & 80.0 & \GapOBQA{5.7} & 89.9 & \GapARCE{6.2} & 83.3 & \GapARCC{5.8} & 63.0 & \GapPIQA{5.1} & 78.9 & \GapAVG{4.3} \\

\midrule
\multicolumn{15}{c}{\textit{End-to-end Systems}} \\
\midrule
Qwen2-Audio-7B     & 71.9 & \GapSC{9.0}  & 29.5 & \GapMMSU{18.7} & 39.6 & \GapOBQA{37.1} & 43.5 & \GapARCE{28.5} & 43.5 & \GapARCC{28.5} & 73.4 & \GapPIQA{5.4} & 50.2 & \GapAVG{21.2} \\
DiVA-Llama3.1-8B   & 82.1 & \GapSC{15.8} & 28.8 & \GapMMSU{34.1} & 40.0 & \GapOBQA{35.0} & 39.7 & \GapARCE{53.6} & 33.7 & \GapARCC{48.9} & 35.6 & \GapPIQA{30.9} & 43.3 & \GapAVG{36.4} \\
GLM-4-Voice-9B     & 76.4 & \GapSC{20.6} & 39.2 & \GapMMSU{28.3} & 52.1 & \GapOBQA{32.5} & 73.2 & \GapARCE{24.6} & 59.5 & \GapARCC{33.8} & 47.3 & \GapPIQA{30.9} & 57.9 & \GapAVG{28.4} \\
Qwen2.5-Omni-7B    & 83.9 & \GapSC{5.4} & 61.3 & \GapMMSU{9.6} &
                      81.5 & \GapOBQA{4.2} & 94.9 & \GapARCE{1.2} &
                      86.1 & \GapARCC{3.0} & 72.8 & \GapPIQA{-4.8} & 80.1 & \GapAVG{3.1} \\
Kimi-Audio-7B         & 66.6 & \GapSC{22.6} & 61.6 & \GapMMSU{9.2} &
                      83.7 & \GapOBQA{2.0} & 95.2 & \GapARCE{0.9} &
                      88.2 & \GapARCC{1.0} & \textbf{82.9} & \GapPIQA{-14.8}[\textbf{-14.8}] & 79.7 & \GapAVG{3.5} \\
\midrule
\multicolumn{15}{c}{\textit{Modality Gap-optimized Systems}} \\
\midrule
\textsc{SALAD}-3B & & & & & & & & & & & & & & & \\
$\quad$Stage~I & 75.5 & \GapSC{7.4} & 47.3 & \GapMMSU{14.6} &
                      65.5 & \GapOBQA{16.3} & - & - &
                      75.6 & \GapARCC{6.2} & 78.3 & \GapPIQA{0.3} & 68.4 & \GapAVG{9.0} \\
$\quad$Stage~II & 75.8 & \GapSC{7.1} & 52.5 & \GapMMSU{9.4} &
                      76.7 & \GapOBQA{5.1} & - & - &
                      79.9 & \GapARCC{1.9} & 78.1 & \GapPIQA{0.5} & 72.6 & \GapAVG{4.8} \\
\textbf{\ourmethod{}-3B}   & 84.2 & \GapSC{8.4} & 57.6 & \GapMMSU{4.0} & 73.4 & \GapOBQA{3.3} & 92.9 & \GapARCE{1.4} & 80.7 & \GapARCC{3.1} & 56.9 & \GapPIQA{-4.8} & 74.3 & \GapAVG{2.6} \\
\textsc{SALAD}-7B & & & & & & & & & & & & & & & \\
$\quad$Stage~I & 81.5 & \GapSC{3.5} & 55.3 & \GapMMSU{15.5} &
                      69.7 & \GapOBQA{19.3} & - & - &
                      82.3 & \GapARCC{6.1} & 80.3 & \GapPIQA{0.4} & 73.8 & \GapAVG{9.0} \\
            
$\quad$Stage~II & 81.5 & \GapSC{3.5} & 57.5 & \GapMMSU{13.3} & 75.1 & \GapOBQA{13.9} & - & - & 84.0 & \GapARCC{4.4} & 80.3 & \GapPIQA{0.4} & 75.7 & \GapAVG{7.1} \\
\textbf{\ourmethod{}-7B} & 88.6 & \GapSC{0.6} & 66.7 & \GapMMSU{4.2} & \textbf{85.1} & \GapOBQA{0.7}[\textbf{0.7}] & 95.8 & \GapARCE{0.3} & \textbf{89.4} & \GapARCC{-0.2}[\textbf{-0.2}] & 69.7 & \GapPIQA{-1.6} & \textbf{82.5} & \GapAVG{0.7}[\textbf{0.7}] \\
\textbf{\ourmethod{}-7B 5:1} & 87.6 & \GapSC{1.6} & 66.2 & \GapMMSU{4.7} & 83.3 & \GapOBQA{2.5} & \textbf{95.9} & \GapARCE{0.2}[\textbf{0.2}] & 88.8 & \GapARCC{0.4} & 70.8 & \GapPIQA{-2.7} & 82.1 & \GapAVG{1.1} \\
\midrule
\multicolumn{15}{c}{\textit{Ablation Studies}} \\
\midrule
\ourmethod{}-7B w/o KD & 84.7 & \GapSC{4.5} & 46.7 & \GapMMSU{24.2} & 55.8 & \GapOBQA{30.0} & 89.5 & \GapARCE{6.6} & 75.8 & \GapARCC{13.4} & 66.6 & \GapPIQA{1.5} & 69.9 & \GapAVG{13.3} \\
\ourencoder{} + Qwen2.5-7B & 88.8 & \GapSC{0.4} & 62.6 & \GapMMSU{8.3} & 80.4 & \GapOBQA{5.4} & 95.0 & \GapARCE{1.1} & 87.8 & \GapARCC{1.4} & 63.3 & \GapPIQA{4.8} & 79.7 & \GapAVG{3.6} \\
\bottomrule
\end{tabular}
    }
    \end{center}
\end{table}

\subsection{LLM Evaluations}

\subsubsection{Experimental Setup}
\textbf{Benchmarks and Evaluations.} We evaluate our LLM on two groups of benchmarks: one for measuring the modality gap and the other for assessing paralinguistic understanding.\footnote{All speech inputs used in our evaluations are shorter than 30 seconds.} To evaluate the modality gap, we follow the benchmark setup of prior work \cite{salad}, including Spoken StoryCloze \cite{twist}, OpenBookQA and MMSU from VoiceBench \cite{voicebench}, PIQA \cite{piqa}, and ARC, including both the Easy and Challenge splits \cite{ai2_arc}. To evaluate paralinguistic understanding, we primarily use the paralinguistic task set from AIR-Bench \cite{airbench}, which covers emotion recognition, speaker gender recognition, and speaker age prediction. All evaluations are conducted in the zero-shot setting.

In addition to the paralinguistic tasks provided by AIR-Bench, we consider speaker accent as another important paralinguistic attribute. Since we did not find a convenient off-the-shelf benchmark for this task, we construct our own accent detection benchmark and evaluate our models on it. Specifically, we use a subset of the Common Voice test set with accent labels and convert it into an AIR-Bench-style four-option multiple-choice benchmark. The four target labels are ``United States'', ``England'', ``India and South Asia (India, Pakistan, Sri Lanka)'', and ``Europe''. We sample 1,000 examples to build the benchmark. Details of the accent benchmark construction are provided in Appendix \ref{apx:accent_benchmark}.
\footnote{We will open-source this accent benchmark.}

\textbf{Baselines.} We compare our model with a diverse set of SLM baselines, including leading industrial systems and methods explicitly designed to reduce the modality gap. We calculate the modality gap as the performance difference between a speech model given spoken input and its backbone TLM given textual input. The compared model pairs are GLM-4-Voice \cite{glm-4-voice} and GLM-4-9B \cite{glm2024chatglm}, Qwen2-Audio \cite{qwen2-audio} and Qwen-7B-Chat \cite{qwen}, DiVA \cite{diva} and Llama-3-8B \cite{llama3}, Qwen2.5-Omni \cite{qwen25-omni} and Qwen2.5-7B-Instruct \cite{qwen2_5}, Kimi-Audio \cite{kimi-audio} and Qwen2.5-7B-Instruct, and SALAD \cite{salad} with Qwen2.5-3B\&7B-Base. As a reference point, we also include cascaded systems that combine Whisper-large-v3 with Qwen2.5-3B\&7B-Instruct.


\subsubsection{Modality Gap Results}
\begin{wraptable}{r}{0.56\textwidth}
\vspace{-1.4em}
\centering
\small
\caption{Reasoning-heavy VoxEval math performance $\uparrow$ and modality gap $\downarrow$ (\%).}
\label{tab:reasoning_modality_gap}
{\setlength{\tabcolsep}{2pt}%
\resizebox{0.56\textwidth}{!}{%
\begin{tabular}{@{}lcccccccc@{}}
\toprule
\multirow{2}{*}{} & \multicolumn{2}{c}{Elementary} & \multicolumn{2}{c}{High School} & \multicolumn{2}{c}{College} & \multicolumn{2}{c}{Avg.} \\
\cmidrule(l){2-9}
& \multicolumn{1}{c}{Acc.} & \multicolumn{1}{c}{Gap}
& \multicolumn{1}{c}{Acc.} & \multicolumn{1}{c}{Gap}
& \multicolumn{1}{c}{Acc.} & \multicolumn{1}{c}{Gap}
& \multicolumn{1}{c}{Acc.} & \multicolumn{1}{c}{Gap} \\
\midrule
\multicolumn{9}{c}{\textit{Text Topline}} \\
\midrule
Qwen2.5-7B-Instruct & 82.1 & \multicolumn{1}{c}{-} & 81.3 & \multicolumn{1}{c}{-} & 80.4 & \multicolumn{1}{c}{-} & 81.3 & \multicolumn{1}{c}{-} \\
\midrule
\multicolumn{9}{c}{\textit{End-to-end Systems}} \\
\midrule
Qwen2.5-Omni-7B & 74.3 & \GapCell{7.85}{20}{0.0}{7.9} & 62.6 & \GapCell{18.71}{20}{0.0}{18.7} & 47.8 & \GapCell{32.60}{20}{0.0}{32.6} & 61.6 & \GapCell{19.72}{20}{0.0}{19.7} \\
Kimi-Audio-7B & 66.1 & \GapCell{16.07}{20}{0.0}{16.1} & 59.7 & \GapCell{21.63}{20}{0.0}{21.6} & 30.4 & \GapCell{50.00}{20}{0.0}{50.0} & 52.1 & \GapCell{29.23}{20}{0.0}{29.2} \\
\textbf{\ourmethod{}-7B} & \textbf{81.4} & \GapCell{0.71}{20}{0.0}{\textbf{0.7}} & \textbf{77.8} & \GapCell{3.50}{20}{0.0}{\textbf{3.5}} & \textbf{69.6} & \GapCell{10.86}{20}{0.0}{\textbf{10.9}} & \textbf{76.3} & \GapCell{5.02}{20}{0.0}{\textbf{5.0}} \\
\bottomrule
\end{tabular}%
}%
}
\end{wraptable}
Following Cuervo \textit{et al.} \cite{salad}, we report both accuracy and modality gap for each dataset and model. The results are summarized in Table~\ref{tab:modality_gap}, where \ourmethod{} denotes the global prepending setting and \ourmethod{} 5:1 denotes the interleaving setting with a 5:1 ratio. We make four main observations.
\textbf{1) \ourmethod{} achieves the lowest average modality gap among all compared SLMs at both the 3B and 7B scales.} In particular, \ourmethod{}-7B attains an average modality gap of 0.7\%, substantially lower than SALAD's 7.1\% and Qwen2.5-Omni's 3.1\%.
\textbf{2) \ourmethod{}-7B 5:1 achieves performance comparable to that of \ourmethod{}-7B.} This result suggests that, even with more injected prosody embeddings, our method preserves semantic capabilities effectively.
\textbf{3) \ourmethod{} consistently achieves a slightly lower modality gap than the cascaded toplines.} On the 7B setting, it yields a lower modality gap on 4 of 6 benchmarks. We suspect that this advantage comes from the dedicated knowledge distillation applied to \ourencoder{}'s outputs, and we examine this hypothesis further in the ablation study. 
\textbf{4) \ourmethod{} shows particularly strong gains on reasoning-required benchmarks compared with SALAD.} Datasets such as MMSU and OBQA require stronger reasoning ability, and \ourmethod{}-7B reduces the gap to 4.2\% and 0.7\%, respectively, compared with SALAD's 13.3\% and 13.9\%.
Overall, these observations suggest that reducing the modality gap from the input side is highly effective, particularly considering \ourmethod{} only requires \textbf{fewer than 300 hours of audio} for knowledge distillation.

\textbf{Ablation Studies.} We study two ablated variants in Table \ref{tab:modality_gap}. \textbf{1) \ourmethod{}-7B w/o KD:} In this setting, we remove knowledge distillation and retain only the paralinguistic-task training. As expected, performance drops substantially. \textbf{2) \ourencoder{} + Qwen2.5-7B:} In this setting, we directly feed the text tokens produced by \ourencoder{} to Qwen2.5-7B-Instruct. We find that performance remains broadly comparable to that of \ourmethod{}, with slight drops on some benchmarks. Because the main differences between this setting and \ourmethod{} are the insertion of prosody embeddings and end-to-end model training, we hypothesize that \ourmethod{} benefits from adapting the LLM to \ourencoder{}-produced texts, making the backbone more compatible with the speech encoder.

\textbf{Modality Gap on Reasoning-Heavy Tasks.}
Although our benchmark choices for measuring the modality gap in Table~\ref{tab:modality_gap} follow prior work, they mostly involve tasks with low to moderate reasoning demands. We therefore further examine the modality gap on more reasoning-heavy benchmarks. Specifically, we use the three math-related tasks in VoxEval \cite{voxeval}: elementary math, high school math, and college math. Because Qwen2.5-Omni-7B and Kimi-Audio-7B show the lowest modality gap among the baseline SLMs in Table~\ref{tab:modality_gap}, we compare these two models with our approach on these three benchmarks. We cap the computational budget at 512 tokens. The results in Table~\ref{tab:reasoning_modality_gap} highlight two main findings. \textbf{1) All SLMs exhibit a larger modality gap on reasoning-heavy tasks.} Specifically, the average modality gap of Qwen2.5-Omni-7B increases from 3.1\% to 19.7\%, while Kimi-Audio-7B shows an average gap above 20\%. \textbf{2) Our approach performs substantially better on reasoning-heavy tasks.} In particular, \ourmethod{} achieves a modality gap of just 3.5\% on high school math, compared with 18.7\% for Qwen2.5-Omni-7B and 21.6\% for Kimi-Audio-7B. Overall, \ourmethod{} also yields the lowest average modality gap across these benchmarks.

\subsubsection{Paralinguistic Understanding Results}
Table \ref{tab:paralinguistic} reports the results on four paralinguistic understanding tasks and reveals two main observations. \textbf{1) \ourmethod{} achieves the strongest paralinguistic understanding performance among all baselines.} This result suggests that even with only a single prosody embedding and roughly 1{,}000 hours of training audio, an SLM can already achieve state-of-the-art paralinguistic understanding. \textbf{2) \ourmethod{}-5:1 performs slightly better than \ourmethod{} on paralinguistic understanding.} In particular, it improves performance on three of the four tasks, likely because feeding more prosody embeddings into the LLM exposes richer acoustic cues. However, the overall gain remains modest, indicating that simply increasing the amount of prosodic input does not substantially improve paralinguistic understanding. Taken together, these results support our hypothesis that, for SLMs, semantic information is more critical than additional acoustic detail.

\begin{wraptable}{r}{0.56\textwidth}
\vspace{-1.2em}
\centering
\caption{SLM performance (\%) on various paralinguistic understanding tasks.}
\label{tab:paralinguistic}
\setlength{\tabcolsep}{2.5pt}
\footnotesize
\begin{tabular*}{\linewidth}{@{\extracolsep{\fill}}lccccc@{}}
\toprule
\textbf{Model} & \textbf{Emo} & \textbf{Gender} & \textbf{Age} & \textbf{Accent} & \textbf{AVG}\\ \midrule
Qwen2-Audio-7B & 48.2 & 64.7 & 23.1 & 29.0 & 41.3 \\
DiVA & 30.7 & 46.8 & 33.3 & 32.5 & 35.8 \\
GLM-4-Voice-9B & 23.0 & 23.9 & 18.7 & 32.4 & 24.5 \\
Qwen2.5-Omni-7B & 54.8 & \textbf{89.8} & 44.8 & 45.3 & 58.7 \\
Kimi-Audio-7B & \textbf{61.1} & 75.9 & 60.5 & 31.2 & 57.2 \\
\textbf{\ourmethod{}-3B} & 58.8 & 80.2 & 63.7 & 45.0 & 61.9 \\
\textbf{\ourmethod{}-7B} & 60.5 & 88.6 & 64.9 & 45.1 & 64.8 \\
\textbf{\ourmethod{}-7B 5:1} & 57.7 & 88.8 & \textbf{66.7} & \textbf{50.2} & \textbf{65.8} \\
\midrule
\multicolumn{6}{c}{\textit{Ablation Studies}} \\
\midrule
\ourmethod{}-7B w/o Recon & 47.3 & 77.7 & 55.5 & 32.2 & 53.2 \\
\ourmethod{}-7B w/o Train & 51.8 & 28.5 & 43.6 & 16.1 & 35.0 \\
\bottomrule
\end{tabular*}
\vspace{-1.0em}
\end{wraptable}
\textbf{Ablation Studies.} We study two ablated variants in Table \ref{tab:paralinguistic}. \textbf{1) Without the reconstruction objective in \ourencoder{} (\ourmethod{}-7B w/o Recon):} In this setting, we use the transcription and prosody embedding from the original Whisper-large-v3, without our reconstruction-based training. Performance drops on all four paralinguistic tasks, indicating that the reconstruction objective is critical for preserving paralinguistic information throughout the pipeline. \textbf{2) Without paralinguistic training (\ourmethod{}-7B w/o Train):} In this setting, we directly feed the prosody embedding from \ourencoder{} into Qwen2.5-7B-Instruct without training the model on paralinguistic tasks. This variant also shows a substantial performance drop, confirming \ourmethod{} effectively acquires paralinguistic information through task-specific training.

\section{Conclusion}
In this work, we propose \ourmethod{}, a speech LLM designed to minimize the modality gap from the input side by making spoken input resemble that of a prosody-aware text LLM. Our approach combines \ourencoder{}, which produces synchronized text tokens and prosody embeddings, with an LLM backbone that preserves the semantic strengths of the original text model while learning to interpret paralinguistic information. Across experiments at both 3B and 7B scales, \ourmethod{} achieves the lowest modality gap among strong SLM baselines and delivers strong performance on paralinguistic understanding benchmarks, while requiring only about 1{,}000 hours of audio training data. These results show that aligning speech inputs with text-style representations, with prosodic information preserved, is a promising and data-efficient direction for building more capable and practical SLMs.

\begin{ack}
The research presented in this paper was partially supported by the Research Grants Council of the Hong Kong Special Administrative Region, China (CUHK 2300246, RGC C1043-24G) and (CUHK 14203425, RGC GRF 2151317).
\end{ack}

\bibliographystyle{unsrtnat}
\bibliography{references}

@inproceedings{SLMsurvey,
  title={Recent advances in speech language models: A survey},
  author={Cui, Wenqian and Yu, Dianzhi and Jiao, Xiaoqi and Meng, Ziqiao and Zhang, Guangyan and Wang, Qichao and Guo, Steven Y and King, Irwin},
  booktitle={Proceedings of the 63rd Annual Meeting of the Association for Computational Linguistics (Volume 1: Long Papers)},
  pages={13943--13970},
  year={2025}
}

@article{llama-omni,
  title={Llama-omni: Seamless speech interaction with large language models},
  author={Fang, Qingkai and Guo, Shoutao and Zhou, Yan and Ma, Zhengrui and Zhang, Shaolei and Feng, Yang},
  journal={arXiv preprint arXiv:2409.06666},
  year={2024}
}

@article{mini-omni,
  title={Mini-omni: Language models can hear, talk while thinking in streaming},
  author={Xie, Zhifei and Wu, Changqiao},
  journal={arXiv preprint arXiv:2408.16725},
  year={2024}
}

@article{moshi,
  title={Moshi: a speech-text foundation model for real-time dialogue},
  author={D{\'e}fossez, Alexandre and Mazar{\'e}, Laurent and Orsini, Manu and Royer, Am{\'e}lie and P{\'e}rez, Patrick and J{\'e}gou, Herv{\'e} and Grave, Edouard and Zeghidour, Neil},
  journal={arXiv preprint arXiv:2410.00037},
  year={2024}
}

@article{turnguide,
  title={Think Before You Talk: Enhancing Meaningful Dialogue Generation in Full-Duplex Speech Language Models with Planning-Inspired Text Guidance},
  author={Cui, Wenqian and Zhu, Lei and Li, Xiaohui and Guo, Zhihan and Bai, Haoli and Hou, Lu and King, Irwin},
  journal={arXiv preprint arXiv:2508.07375},
  year={2025}
}

@article{gpt4,
  title={Gpt-4 technical report},
  author={Achiam, Josh and Adler, Steven and Agarwal, Sandhini and Ahmad, Lama and Akkaya, Ilge and Aleman, Florencia Leoni and Almeida, Diogo and Altenschmidt, Janko and Altman, Sam and Anadkat, Shyamal and others},
  journal={arXiv preprint arXiv:2303.08774},
  year={2023}
}

@article{gpt5,
  title={Openai gpt-5 system card},
  author={Singh, Aaditya and Fry, Adam and Perelman, Adam and Tart, Adam and Ganesh, Adi and El-Kishky, Ahmed and McLaughlin, Aidan and Low, Aiden and Ostrow, AJ and Ananthram, Akhila and others},
  journal={arXiv preprint arXiv:2601.03267},
  year={2025}
}

@article{llama4,
  title={The Llama 4 Herd: Architecture, Training, Evaluation, and Deployment Notes},
  author={Adcock, Aaron and Srivastava, Aayushi and Dubey, Abhimanyu and Jauhri, Abhinav and Pande, Abhinav and Pandey, Abhinav and Sharma, Abhinav and Kadian, Abhishek and Kumawat, Abhishek and Kelsey, Adam and others},
  journal={arXiv preprint arXiv:2601.11659},
  year={2026}
}

@article{glm2024chatglm,
  title={Chatglm: A family of large language models from glm-130b to glm-4 all tools},
  author={Glm, Team and Zeng, Aohan and Xu, Bin and Wang, Bowen and Zhang, Chenhui and Yin, Da and Zhang, Dan and Rojas, Diego and Feng, Guanyu and Zhao, Hanlin and others},
  journal={arXiv preprint arXiv:2406.12793},
  year={2024}
}

@article{gemini,
  title={Gemini: a family of highly capable multimodal models},
  author={Team, Gemini and Anil, Rohan and Borgeaud, Sebastian and Alayrac, Jean-Baptiste and Yu, Jiahui and Soricut, Radu and Schalkwyk, Johan and Dai, Andrew M and Hauth, Anja and Millican, Katie and others},
  journal={arXiv preprint arXiv:2312.11805},
  year={2023}
}

@article{gemini1_5,
  title={Gemini 1.5: Unlocking multimodal understanding across millions of tokens of context},
  author={Team, Gemini and Georgiev, Petko and Lei, Ving Ian and Burnell, Ryan and Bai, Libin and Gulati, Anmol and Tanzer, Garrett and Vincent, Damien and Pan, Zhufeng and Wang, Shibo and others},
  journal={arXiv preprint arXiv:2403.05530},
  year={2024}
}

@article{glm-4-voice,
  title={Glm-4-voice: Towards intelligent and human-like end-to-end spoken chatbot},
  author={Zeng, Aohan and Du, Zhengxiao and Liu, Mingdao and Wang, Kedong and Jiang, Shengmin and Zhao, Lei and Dong, Yuxiao and Tang, Jie},
  journal={arXiv preprint arXiv:2412.02612},
  year={2024}
}

@article{salad,
  title={Closing the gap between text and speech understanding in llms},
  author={Cuervo, Santiago and Seto, Skyler and de Seyssel, Maureen and Bai, Richard He and Gu, Zijin and Likhomanenko, Tatiana and Jaitly, Navdeep and Aldeneh, Zakaria},
  journal={arXiv preprint arXiv:2510.13632},
  year={2025}
}

@article{twist,
  title={Textually pretrained speech language models},
  author={Hassid, Michael and Remez, Tal and Nguyen, Tu Anh and Gat, Itai and Conneau, Alexis and Kreuk, Felix and Copet, Jade and Defossez, Alexandre and Synnaeve, Gabriel and Dupoux, Emmanuel and others},
  journal={Advances in Neural Information Processing Systems},
  volume={36},
  year={2024}
}

@article{spiritlm,
  title={Spirit-lm: Interleaved spoken and written language model},
  author={Nguyen, Tu Anh and Muller, Benjamin and Yu, Bokai and Costa-Jussa, Marta R and Elbayad, Maha and Popuri, Sravya and Duquenne, Paul-Ambroise and Algayres, Robin and Mavlyutov, Ruslan and Gat, Itai and others},
  journal={arXiv preprint arXiv:2402.05755},
  year={2024}
}

@article{OnGenerativeRawAudio,
  title={On generative spoken language modeling from raw audio},
  author={Lakhotia, Kushal and Kharitonov, Eugene and Hsu, Wei-Ning and Adi, Yossi and Polyak, Adam and Bolte, Benjamin and Nguyen, Tu-Anh and Copet, Jade and Baevski, Alexei and Mohamed, Abdelrahman and others},
  journal={Transactions of the Association for Computational Linguistics},
  volume={9},
  pages={1336--1354},
  year={2021},
  publisher={MIT Press One Rogers Street, Cambridge, MA 02142-1209, USA journals-info~…}
}

@inproceedings{speechgpt,
    title = "{S}peech{GPT}: Empowering Large Language Models with Intrinsic Cross-Modal Conversational Abilities",
    author = "Zhang, Dong  and
      Li, Shimin  and
      Zhang, Xin  and
      Zhan, Jun  and
      Wang, Pengyu  and
      Zhou, Yaqian  and
      Qiu, Xipeng",
    editor = "Bouamor, Houda  and
      Pino, Juan  and
      Bali, Kalika",
    booktitle = "Findings of the Association for Computational Linguistics: EMNLP 2023",
    month = dec,
    year = "2023",
    address = "Singapore",
    publisher = "Association for Computational Linguistics",
    url = "https://aclanthology.org/2023.findings-emnlp.1055",
    doi = "10.18653/v1/2023.findings-emnlp.1055",
    pages = "15757--15773",
}

@inproceedings{vita-audio,
  title={VITA-Audio: Fast Interleaved Audio-Text Token Generation for Efficient Large Speech-Language Model},
  author={Long, Zuwei and Shen, Yunhang and Fu, Chaoyou and Gao, Heting and Chen, Peixian and Zhang, Mengdan and Shao, Hang and Li, Jian and Peng, Jinlong and Cao, Haoyu and others},
  year = {2025},
  booktitle={The Thirty-ninth Annual Conference on Neural Information Processing Systems}
}

@inproceedings{llama-omni2,
  title={LLaMA-omni 2: LLM-based real-time spoken chatbot with autoregressive streaming speech synthesis},
  author={Fang, Qingkai and Zhou, Yan and Guo, Shoutao and Zhang, Shaolei and Feng, Yang},
  booktitle={Proceedings of the 63rd Annual Meeting of the Association for Computational Linguistics (Volume 1: Long Papers)},
  pages={18617--18629},
  year={2025}
}

@article{minmo,
  title={Minmo: A multimodal large language model for seamless voice interaction},
  author={Chen, Qian and Chen, Yafeng and Chen, Yanni and Chen, Mengzhe and Chen, Yingda and Deng, Chong and Du, Zhihao and Gao, Ruize and Gao, Changfeng and Gao, Zhifu and others},
  journal={arXiv preprint arXiv:2501.06282},
  year={2025}
}

@inproceedings{vocalnet,
  title={VocalNet: Speech LLMs with Multi-Token Prediction for Faster and High-Quality Generation},
  author={Wang, Yuhao and Liu, Heyang and Cheng, Ziyang and Wu, Ronghua and Gu, Qunshan and Wang, Yanfeng and Wang, Yu},
  booktitle={Proceedings of the 2025 Conference on Empirical Methods in Natural Language Processing},
  pages={19595--19612},
  year={2025}
}

@article{qwen3-omni,
  title={Qwen3-omni technical report},
  author={Xu, Jin and Guo, Zhifang and Hu, Hangrui and Chu, Yunfei and Wang, Xiong and He, Jinzheng and Wang, Yuxuan and Shi, Xian and He, Ting and Zhu, Xinfa and others},
  journal={arXiv preprint arXiv:2509.17765},
  year={2025}
}

@misc{qwen25-omni,
      title={Qwen2.5-Omni Technical Report}, 
      author={Jin Xu and Zhifang Guo and Jinzheng He and Hangrui Hu and Ting He and Shuai Bai and Keqin Chen and Jialin Wang and Yang Fan and Kai Dang and Bin Zhang and Xiong Wang and Yunfei Chu and Junyang Lin},
      year={2025},
      eprint={2503.20215},
      archivePrefix={arXiv},
      primaryClass={cs.CL},
      url={https://arxiv.org/abs/2503.20215}, 
}

@article{wav2vec20,
  title={wav2vec 2.0: A framework for self-supervised learning of speech representations},
  author={Baevski, Alexei and Zhou, Yuhao and Mohamed, Abdelrahman and Auli, Michael},
  journal={Advances in neural information processing systems},
  volume={33},
  pages={12449--12460},
  year={2020}
}

@inproceedings{w2v-bert,
  title={W2v-bert: Combining contrastive learning and masked language modeling for self-supervised speech pre-training},
  author={Chung, Yu-An and Zhang, Yu and Han, Wei and Chiu, Chung-Cheng and Qin, James and Pang, Ruoming and Wu, Yonghui},
  booktitle={2021 IEEE Automatic Speech Recognition and Understanding Workshop (ASRU)},
  pages={244--250},
  year={2021},
  organization={IEEE}
}

@article{hubert,
  title={Hubert: Self-supervised speech representation learning by masked prediction of hidden units},
  author={Hsu, Wei-Ning and Bolte, Benjamin and Tsai, Yao-Hung Hubert and Lakhotia, Kushal and Salakhutdinov, Ruslan and Mohamed, Abdelrahman},
  journal={IEEE/ACM transactions on audio, speech, and language processing},
  volume={29},
  pages={3451--3460},
  year={2021},
  publisher={IEEE}
}

@article{kimi-audio,
  title={Kimi-audio technical report},
  author={Ding, Ding and Ju, Zeqian and Leng, Yichong and Liu, Songxiang and Liu, Tong and Shang, Zeyu and Shen, Kai and Song, Wei and Tan, Xu and Tang, Heyi and others},
  journal={arXiv preprint arXiv:2504.18425},
  year={2025}
}

@article{layerwise_distill,
  title={Teaching Audio Models to Reason: A Unified Framework for Source-and Layer-wise Distillation},
  author={Yang, Runyan and Si, Yuke and Gao, Yingying and Feng, Junlan and Deng, Chao and Zhang, Shilei},
  journal={arXiv preprint arXiv:2509.18579},
  year={2025}
}

@article{cross_modal_distillation,
  title={Cross-modal knowledge distillation for speech large language models},
  author={Wang, Enzhi and Li, Qicheng and Tang, Zhiyuan and Jia, Yuhang},
  journal={arXiv preprint arXiv:2509.14930},
  year={2025}
}

@article{cord,
  title={CORD: Bridging the Audio-Text Reasoning Gap via Weighted On-policy Cross-modal Distillation},
  author={Hu, Jing and Zhu, Danxiang and Luo, Xianlong and Zhang, Dan and He, Shuwei and Lei, Yishu and Zheng, Haitao and Feng, Shikun and He, Jingzhou and Sun, Yu and others},
  journal={arXiv preprint arXiv:2601.16547},
  year={2026}
}

@article{deepomni,
  title={DeepOmni: Towards Seamless and Smart Speech Interaction with Adaptive Modality-Specific MoE},
  author={Shao, Hang and Gao, Heting and Shen, Yunhang and Chen, Jiawei and Long, Zuwei and Yang, Dong and Li, Ke and Sun, Xing},
  journal={arXiv preprint arXiv:2506.21864},
  year={2025}
}

@article{gap_continual_learning,
  title={Analyzing mitigation strategies for catastrophic forgetting in end-to-end training of spoken language models},
  author={Hsiao, Chi-Yuan and Lu, Ke-Han and Chang, Kai-Wei and Yang, Chih-Kai and Chen, Wei-Chih and Lee, Hung-yi},
  journal={arXiv preprint arXiv:2505.17496},
  year={2025}
}

@article{anatomy_gap_understanding_long_parse,
  title={Anatomy of the Modality Gap: Dissecting the Internal States of End-to-End Speech LLMs},
  author={Hsu, Ming-Hao and Zhang, Xueyao and Tian, Xiaohai and Zhang, Jun and Wu, Zhizheng},
  journal={arXiv preprint arXiv:2603.01502},
  year={2026}
}

@article{x-opd,
  title={X-OPD: Cross-Modal On-Policy Distillation for Capability Alignment in Speech LLMs},
  author={Cao, Di and Fu, Dongjie and Yu, Hai and Zheng, Siqi and Tan, Xu and Jin, Tao},
  journal={arXiv preprint arXiv:2603.24596},
  year={2026}
}

@inproceedings{gap_understanding_direction_magnitude,
  title={Understanding the Modality Gap: An Empirical Study on the Speech-Text Alignment Mechanism of Large Speech Language Models},
  author={Xiang, Bajian and Zhao, Shuaijiang and Guo, Tingwei and Zou, Wei},
  booktitle={Proceedings of the 2025 Conference on Empirical Methods in Natural Language Processing},
  pages={5187--5202},
  year={2025}
}

@article{MoE_survey,
  title={A comprehensive survey of mixture-of-experts: Algorithms, theory, and applications},
  author={Mu, Siyuan and Lin, Sen},
  journal={arXiv preprint arXiv:2503.07137},
  year={2025}
}

@article{multimodal_continual_learning_survey,
  title={Recent advances of multimodal continual learning: A comprehensive survey},
  author={Yu, Dianzhi and Zhang, Xinni and Chen, Yankai and Liu, Aiwei and Zhang, Yifei and Yu, Philip S and King, Irwin},
  journal={arXiv preprint arXiv:2410.05352},
  year={2024}
}

@inproceedings{whisper,
  title={Robust speech recognition via large-scale weak supervision},
  author={Radford, Alec and Kim, Jong Wook and Xu, Tao and Brockman, Greg and McLeavey, Christine and Sutskever, Ilya},
  booktitle={International conference on machine learning},
  pages={28492--28518},
  year={2023},
  organization={PMLR}
}

@article{wavtokenizer,
  title={Wavtokenizer: an efficient acoustic discrete codec tokenizer for audio language modeling},
  author={Ji, Shengpeng and Jiang, Ziyue and Wang, Wen and Chen, Yifu and Fang, Minghui and Zuo, Jialong and Yang, Qian and Cheng, Xize and Wang, Zehan and Li, Ruiqi and others},
  journal={arXiv preprint arXiv:2408.16532},
  year={2024}
}

@inproceedings{librispeech,
  title={Librispeech: an asr corpus based on public domain audio books},
  author={Panayotov, Vassil and Chen, Guoguo and Povey, Daniel and Khudanpur, Sanjeev},
  booktitle={2015 IEEE international conference on acoustics, speech and signal processing (ICASSP)},
  pages={5206--5210},
  year={2015},
  organization={IEEE}
}

@inproceedings{librispeech-pc,
  title={Librispeech-pc: Benchmark for evaluation of punctuation and capitalization capabilities of end-to-end asr models},
  author={Meister, Aleksandr and Novikov, Matvei and Karpov, Nikolay and Bakhturina, Evelina and Lavrukhin, Vitaly and Ginsburg, Boris},
  booktitle={2023 IEEE automatic speech recognition and understanding workshop (ASRU)},
  pages={1--7},
  year={2023},
  organization={IEEE}
}

@inproceedings{commonsenseqa,
  title={Commonsenseqa: A question answering challenge targeting commonsense knowledge},
  author={Talmor, Alon and Herzig, Jonathan and Lourie, Nicholas and Berant, Jonathan},
  booktitle={Proceedings of the 2019 Conference of the North American Chapter of the Association for Computational Linguistics: Human Language Technologies, Volume 1 (Long and Short Papers)},
  pages={4149--4158},
  year={2019}
}

@inproceedings{ultrachat,
  title={Enhancing chat language models by scaling high-quality instructional conversations},
  author={Ding, Ning and Chen, Yulin and Xu, Bokai and Qin, Yujia and Hu, Shengding and Liu, Zhiyuan and Sun, Maosong and Zhou, Bowen},
  booktitle={Proceedings of the 2023 Conference on Empirical Methods in Natural Language Processing},
  pages={3029--3051},
  year={2023}
}

@inproceedings{paraspeechcaps,
  title={Scaling rich style-prompted text-to-speech datasets},
  author={Diwan, Anuj and Zheng, Zhisheng and Harwath, David and Choi, Eunsol},
  booktitle={Proceedings of the 2025 Conference on Empirical Methods in Natural Language Processing},
  pages={3639--3659},
  year={2025}
}

@article{iemocap,
  title={IEMOCAP: Interactive emotional dyadic motion capture database},
  author={Busso, Carlos and Bulut, Murtaza and Lee, Chi-Chun and Kazemzadeh, Abe and Mower, Emily and Kim, Samuel and Chang, Jeannette N and Lee, Sungbok and Narayanan, Shrikanth S},
  journal={Language resources and evaluation},
  volume={42},
  number={4},
  pages={335--359},
  year={2008},
  publisher={Springer}
}

@misc{crema-d,
  title={Crowd-sourced Emotional Multimodal Actors Dataset (CREMA-D)},
  author={Cao, Houwei and Cooper, David G and Keutmann, Michael K and Gur, Ruben C and Nenkova, Ani and Verma, Ragini},
  year={2025}
}

@book{tess,
  title={Toronto emotional speech set (TESS)},
  author={Dupuis, Kate and Pichora-Fuller, M Kathleen},
  year={2010},
  publisher={University of Toronto, Psychology Department Toronto, ON, Canada}
}

@article{savee,
  title={Surrey audio-visual expressed emotion (savee) database},
  author={Jackson, Philip and Haq, SJUoSG},
  journal={University of Surrey: Guildford, UK},
  year={2014}
}

@article{ravdess,
  title={The Ryerson Audio-Visual Database of Emotional Speech and Song (RAVDESS): A dynamic, multimodal set of facial and vocal expressions in North American English},
  author={Livingstone, Steven R and Russo, Frank A},
  journal={PloS one},
  volume={13},
  number={5},
  pages={e0196391},
  year={2018},
  publisher={Public Library of Science}
}

@article{esd,
  title={Emotional voice conversion: Theory, databases and esd},
  author={Zhou, Kun and Sisman, Berrak and Liu, Rui and Li, Haizhou},
  journal={Speech Communication},
  volume={137},
  pages={1--18},
  year={2022},
  publisher={Elsevier}
}

@inproceedings{commonvoice,
  title={Common voice: A massively-multilingual speech corpus},
  author={Ardila, Rosana and Branson, Megan and Davis, Kelly and Kohler, Michael and Meyer, Josh and Henretty, Michael and Morais, Reuben and Saunders, Lindsay and Tyers, Francis and Weber, Gregor},
  booktitle={Proceedings of the twelfth language resources and evaluation conference},
  pages={4218--4222},
  year={2020}
}

@inproceedings{airbench,
  title={Air-bench: Benchmarking large audio-language models via generative comprehension},
  author={Yang, Qian and Xu, Jin and Liu, Wenrui and Chu, Yunfei and Jiang, Ziyue and Zhou, Xiaohuan and Leng, Yichong and Lv, Yuanjun and Zhao, Zhou and Zhou, Chang and others},
  booktitle={Proceedings of the 62nd Annual Meeting of the Association for Computational Linguistics (Volume 1: Long Papers)},
  pages={1979--1998},
  year={2024}
}

@article{voicebench,
  title={Voicebench: Benchmarking llm-based voice assistants},
  author={Chen, Yiming and Yue, Xianghu and Zhang, Chen and Gao, Xiaoxue and Tan, Robby T and Li, Haizhou},
  journal={Transactions of the Association for Computational Linguistics},
  volume={14},
  pages={378--398},
  year={2026},
  publisher={MIT Press 255 Main Street, 9th Floor, Cambridge, Massachusetts 02142, USA~…}
}

@inproceedings{piqa,
  title={Piqa: Reasoning about physical commonsense in natural language},
  author={Bisk, Yonatan and Zellers, Rowan and Gao, Jianfeng and Choi, Yejin and others},
  booktitle={Proceedings of the AAAI conference on artificial intelligence},
  volume={34},
  pages={7432--7439},
  year={2020}
}

@article{ai2_arc,
  title={Think you have solved question answering? try arc, the ai2 reasoning challenge},
  author={Clark, Peter and Cowhey, Isaac and Etzioni, Oren and Khot, Tushar and Sabharwal, Ashish and Schoenick, Carissa and Tafjord, Oyvind},
  journal={arXiv preprint arXiv:1803.05457},
  year={2018}
}

@article{qwen2-audio,
  title={Qwen2-audio technical report},
  author={Chu, Yunfei and Xu, Jin and Yang, Qian and Wei, Haojie and Wei, Xipin and Guo, Zhifang and Leng, Yichong and Lv, Yuanjun and He, Jinzheng and Lin, Junyang and others},
  journal={arXiv preprint arXiv:2407.10759},
  year={2024}
}

@inproceedings{diva,
  title={Distilling an end-to-end voice assistant without instruction training data},
  author={Held, William and Zhang, Yanzhe and Li, Minzhi and Shi, Weiyan and Ryan, Michael J and Yang, Diyi},
  booktitle={Proceedings of the 63rd Annual Meeting of the Association for Computational Linguistics (Volume 1: Long Papers)},
  pages={7876--7891},
  year={2025}
}

@article{gigaspeech,
  title={Gigaspeech: An evolving, multi-domain asr corpus with 10,000 hours of transcribed audio},
  author={Chen, Guoguo and Chai, Shuzhou and Wang, Guanbo and Du, Jiayu and Zhang, Wei-Qiang and Weng, Chao and Su, Dan and Povey, Daniel and Trmal, Jan and Zhang, Junbo and others},
  journal={arXiv preprint arXiv:2106.06909},
  year={2021}
}

@article{voxceleb1,
  title={Voxceleb: Large-scale speaker verification in the wild},
  author={Nagrani, Arsha and Chung, Joon Son and Xie, Weidi and Zisserman, Andrew},
  journal={Computer Speech \& Language},
  volume={60},
  pages={101027},
  year={2020},
  publisher={Elsevier}
}

@article{libritts,
  title={Libritts: A corpus derived from librispeech for text-to-speech},
  author={Zen, Heiga and Dang, Viet and Clark, Rob and Zhang, Yu and Weiss, Ron J and Jia, Ye and Chen, Zhifeng and Wu, Yonghui},
  journal={arXiv preprint arXiv:1904.02882},
  year={2019}
}

@article{llama3,
  title={The llama 3 herd of models},
  author={Grattafiori, Aaron and Dubey, Abhimanyu and Jauhri, Abhinav and Pandey, Abhinav and Kadian, Abhishek and Al-Dahle, Ahmad and Letman, Aiesha and Mathur, Akhil and Schelten, Alan and Vaughan, Alex and others},
  journal={arXiv preprint arXiv:2407.21783},
  year={2024}
}

@article{qwen,
  title={Qwen technical report},
  author={Bai, Jinze and Bai, Shuai and Chu, Yunfei and Cui, Zeyu and Dang, Kai and Deng, Xiaodong and Fan, Yang and Ge, Wenbin and Han, Yu and Huang, Fei and others},
  journal={arXiv preprint arXiv:2309.16609},
  year={2023}
}

@misc{qwen2_5,
      title={Qwen2.5 Technical Report}, 
      author={Qwen Team},
      year={2024},
      eprint={2412.15115},
      archivePrefix={arXiv},
      primaryClass={cs.CL},
      url={https://arxiv.org/abs/2412.15115}, 
}

@inproceedings{soundwave,
  title={Soundwave: Less is more for speech-text alignment in llms},
  author={Zhang, Yuhao and Liu, Zhiheng and Bu, Fan and Zhang, Ruiyu and Wang, Benyou and Li, Haizhou},
  booktitle={Proceedings of the 63rd Annual Meeting of the Association for Computational Linguistics (Volume 1: Long Papers)},
  pages={18718--18738},
  year={2025}
}

@article{audiolm,
  title={Audiolm: a language modeling approach to audio generation},
  author={Borsos, Zal{\'a}n and Marinier, Rapha{\"e}l and Vincent, Damien and Kharitonov, Eugene and Pietquin, Olivier and Sharifi, Matt and Roblek, Dominik and Teboul, Olivier and Grangier, David and Tagliasacchi, Marco and others},
  journal={IEEE/ACM transactions on audio, speech, and language processing},
  volume={31},
  pages={2523--2533},
  year={2023},
  publisher={IEEE}
}

@inproceedings{pgslm,
  title={Text-free prosody-aware generative spoken language modeling},
  author={Kharitonov, Eugene and Lee, Ann and Polyak, Adam and Adi, Yossi and Copet, Jade and Lakhotia, Kushal and Nguyen, Tu-Anh and Riviere, Morgane and Mohamed, Abdelrahman and Dupoux, Emmanuel and others},
  booktitle={Proceedings of the 60th Annual Meeting of the Association for Computational Linguistics (Volume 1: Long Papers)},
  pages={8666--8681},
  year={2022}
}

@article{naturalspeech3,
  title={Naturalspeech 3: Zero-shot speech synthesis with factorized codec and diffusion models},
  author={Ju, Zeqian and Wang, Yuancheng and Shen, Kai and Tan, Xu and Xin, Detai and Yang, Dongchao and Liu, Yanqing and Leng, Yichong and Song, Kaitao and Tang, Siliang and others},
  journal={arXiv preprint arXiv:2403.03100},
  year={2024}
}

@article{prosodylm,
  title={ProsodyLM: Uncovering the Emerging Prosody Processing Capabilities in Speech Language Models},
  author={Qian, Kaizhi and Fan, Xulin and Ni, Junrui and Shechtman, Slava and Hasegawa-Johnson, Mark and Gan, Chuang and Zhang, Yang},
  journal={arXiv preprint arXiv:2507.20091},
  year={2025}
}

@article{speechgpt-gen,
  title={Speechgpt-gen: Scaling chain-of-information speech generation},
  author={Zhang, Dong and Zhang, Xin and Zhan, Jun and Li, Shimin and Zhou, Yaqian and Qiu, Xipeng},
  journal={arXiv preprint arXiv:2401.13527},
  year={2024}
}

@article{hussein2025hasrd,
  title={HASRD: Hierarchical Acoustic and Semantic Representation Disentanglement},
  author={Hussein, Amir and Khurana, Sameer and Wichern, Gordon and Germain, Francois G and Roux, Jonathan Le},
  journal={arXiv preprint arXiv:2506.00843},
  year={2025}
}

@inproceedings{voxeval,
  title={Voxeval: Benchmarking the knowledge understanding capabilities of end-to-end spoken language models},
  author={Cui, Wenqian and Jiao, Xiaoqi and Meng, Ziqiao and King, Irwin},
  booktitle={Proceedings of the 63rd Annual Meeting of the Association for Computational Linguistics (Volume 1: Long Papers)},
  pages={16735--16753},
  year={2025}
}

@inproceedings{drvoice,
  title={Drvoice: Parallel speech-text voice conversation model via dual-resolution speech representations},
  author={Tan, Chao-Hong and Chen, Qian and Wang, Wen and Deng, Chong and Zhang, Qinglin and Cheng, Luyao and Yu, Hai and Zhang, Xin and Lyu, Xiang and Zhao, Tianyu and others},
  booktitle={International Conference on Learning Representations},
  volume={2026},
  pages={121421--121440},
  year={2026}
}


\appendix

\section{Discussions \& Limitations}
\label{apx:limitations}
Although we have shown promising results for \ourmethod{}, the current study has several limitations. First, we do not train or evaluate a speech synthesis module, and all comparisons are made on text outputs. We make this choice deliberately because modality-gap studies in recent Speech LLM literature typically compare benchmark performance using text outputs, which has become a standard evaluation protocol. Under the increasingly common decoupled ``thinking'' and ``talking'' design, we view text generation as the component that most directly reflects the model's intelligence and understanding ability, while speech synthesis primarily concerns how accurately and naturally the generated text is pronounced. These two aspects are usually measured with different metrics, such as task accuracy for text generation and metrics such as WER and UTMOS for speech synthesis. 

Second, although representative, our paralinguistic evaluation does not reflect the whole paralinguistic space. We choose a set of tasks that covers several important paralinguistic phenomena and use them to test whether the model can exploit prosodic information beyond plain lexical content. However, real-world spoken interaction may require understanding more complex and dynamic signals, such as abrupt emotion shifts, mixed affective states, speaker intention changes, or other nuanced discourse-level cues. Therefore, our paralinguistic results should be interpreted as a proof of concept that \ourmethod{} can achieve strong paralinguistic understanding rather than achieving state-of-the-art performance on the whole paralinguistic space.

Third, our current implementation may introduce additional latency. In our setup, the speech encoder runs only after the user has finished speaking, because the Whisper backbone we use does not support streaming inference. As a result, the system is sequential rather than fully streamable, which can slightly increase response latency in interactive scenarios. We view this issue primarily as a limitation of the underlying Whisper backbone rather than of the core idea of our method. In principle, the same input-side modality-gap reduction strategy could be combined with a streaming ASR backbone to support lower-latency processing. We leave this direction to future work.

Fourth, the current implementation of \ourmethod{} only supports speech inputs up to 30 seconds long. This restriction stems from the 30-second audio context window of the Whisper backbone used by \ourencoder{} \cite{whisper}. We expect that training \ourencoder{} with an alternative ASR backbone that supports a longer audio context could address this limitation and enable \ourmethod{} to process longer speech inputs.

Finally, adapting our current framework to non-speech audio is non-trivial. Our current pipeline explicitly decomposes the input into a text transcription stream and a prosody embedding stream, which is well-suited to spoken language but not to general audio events. In more complex settings, a model may need to interpret non-speech sounds or mixed audio inputs that cannot be naturally represented through standard transcription. Extending the framework to such cases will likely require dedicated training data and new modeling techniques. One possible direction is to let the text stream emit placeholder tokens, such as masked tokens, when the input contains non-speech audio, while relying on the prosody or acoustic embedding stream to preserve the relevant information. We consider this extension important but outside the scope of the present paper.

\section{Detailed Architecture Explanation of \ourencoder{}}
\label{apx:mel-reconstructor_details}
\paragraph{Input and Whisper backbone.}
The architecture of \ourencoder{} includes two main components: Whisper-large-v3 (encoder) and Mel Reconstructor (decoder). Let an input utterance be a raw waveform \(\mathbf{x}\in\mathbb{R}^{T}\), where \(T\) denotes the number of waveform samples. Following Whisper's standard preprocessing pipeline, the waveform is first resampled and normalized to a fixed \(30\)-second window, and then converted into a log-Mel spectrogram $\mathbf{M}\in\mathbb{R}^{F\times L}$, where \(F=128\) is the number of Mel bins and \(L=3000\) is the number of time frames. Thus, regardless of the original utterance duration, the acoustic input to \ourencoder{} is always represented as a \(128\times 3000\) Mel spectrogram.

Whisper-large-v3 then processes \(\mathbf{M}\) using its encoder-decoder Transformer architecture. First, the Whisper encoder maps the acoustic input into a sequence of latent speech representations,
\begin{equation}
\mathbf{H}^{\mathrm{enc}}=\mathrm{Enc}_{\mathrm{whisper}}(\mathbf{M})
\in\mathbb{R}^{L_e\times d},
\label{eq:whisper-enc}
\end{equation}
where \(L_e\) denotes the encoder sequence length and \(d\) is the Whisper hidden dimension. Conditioned on \(\mathbf{H}^{\mathrm{enc}}\), the Whisper decoder autoregressively predicts a text token sequence
\begin{equation}
\mathbf{y}=(y_1,\dots,y_N), \qquad y_i\in\mathcal{V},
\label{eq:whisper-decoder-output}
\end{equation}
with \(\mathcal{V}\) the text vocabulary and \(N\) the output token length. Concretely, at decoder layer \(l\), the hidden states are represented as
\begin{equation}
\mathbf{H}^{(l)}=(\mathbf{h}^{(l)}_1,\dots,\mathbf{h}^{(l)}_N)\in\mathbb{R}^{N\times d}.
\label{eq:whisper-layer-hidden}
\end{equation}
In our model, the prosody embeddings are taken from the \(16\)-th Whisper decoder layer:
\begin{equation}
\mathbf{P}=\mathbf{H}^{(16)}\in\mathbb{R}^{N\times d}.
\label{eq:prosody-repr}
\end{equation}
Because \(\mathbf{P}\) is extracted from the decoder states aligned with the generated text sequence, each prosody vector \(\mathbf{p}_i\) is naturally paired with exactly one text token \(y_i\). Therefore, the model represents spoken input as two synchronized streams:
\begin{equation}
\{(y_i,\mathbf{p}_i)\}_{i=1}^{N},
\label{eq:sync-streams}
\end{equation}
where \(y_i\) captures the linguistic content and \(\mathbf{p}_i\) captures the associated prosodic information.

\paragraph{Mel reconstructor.}
The Mel Reconstructor is designed to recover the original Mel spectrogram \(\mathbf{M}\) from these two token-aligned streams. Its role is to force the decoder hidden states to preserve information about how the utterance was spoken, rather than only what was said. To this end, each text token \(y_i\) is first embedded into a continuous vector,
\begin{equation}
\mathbf{e}_i=\mathrm{Emb}(y_i)\in\mathbb{R}^{d_r},
\label{eq:text-embedding}
\end{equation}
where \(d_r\) is the hidden size of the reconstructor. In parallel, each prosody embedding \(\mathbf{p}_i\in\mathbb{R}^{d}\) is projected into the same space:
\begin{equation}
\tilde{\mathbf{p}}_i=\mathbf{W}_p\mathbf{p}_i+\mathbf{b}_p \in\mathbb{R}^{d_r}.
\label{eq:prosody-projection-recon}
\end{equation}
The two representations are then fused token-wise. Specifically, they are concatenated first and then projected by a feed-forward layer:
\begin{equation}
\mathbf{z}_i=\phi\!\left([\mathbf{e}_i;\tilde{\mathbf{p}}_i]\right)\in\mathbb{R}^{d_r},
\label{eq:fusion}
\end{equation}
where \([\cdot;\cdot]\) denotes concatenation and \(\phi(\cdot)\) denotes a fusion block consisting of a linear layer, layer normalization, GELU activation, and dropout. Stacking the resulting token-level fused vectors yields
\begin{equation}
\mathbf{Z}=(\mathbf{z}_1,\dots,\mathbf{z}_N)\in\mathbb{R}^{N\times d_r}.
\label{eq:fused-sequence}
\end{equation}

To encode contextual dependencies across the whole utterance, \(\mathbf{Z}\) is augmented with sinusoidal positional encoding and passed through a Transformer encoder:
\begin{equation}
\mathbf{U}=\mathrm{Enc}_{\mathrm{mel}}(\mathbf{Z})\in\mathbb{R}^{N\times d_r}.
\label{eq:mel-encoder}
\end{equation}
This encoded sequence \(\mathbf{U}\) serves as the memory for Mel reconstruction. Intuitively, \(\mathbf{U}\) integrates token identity and prosodic realization into a contextualized latent sequence.

The decoder side of the Mel Reconstructor is implemented as a Transformer decoder with cross-attention and a fixed set of learnable frame queries, one for each target Mel frame. This design addresses the mismatch between the variable-length token-level input sequence and the fixed-length frame-level output sequence produced by Whisper's preprocessing. Rather than imposing a one-to-one alignment between tokens and Mel frames, cross-attention allows each frame query to attend over the full token-level memory and extract the information needed to reconstruct its corresponding acoustic frame. Specifically, we introduce
\begin{equation}
\mathbf{Q}=(\mathbf{q}_1,\dots,\mathbf{q}_{L})\in\mathbb{R}^{L\times d_r},
\label{eq:frame-queries}
\end{equation}
where \(L=3000\) matches the number of Mel frames in Whisper's normalized input. We first add the same sinusoidal positional encoding to \(\mathbf{Q}\) and feed it into a Transformer decoder that performs cross-attention over the memory \(\mathbf{U}\):
\begin{equation}
\mathbf{D}=\mathrm{Dec}_{\mathrm{mel}}(\mathbf{Q},\mathbf{U})\in\mathbb{R}^{L\times d_r}.
\label{eq:mel-decoder}
\end{equation}
Here, each query \(\mathbf{q}_t\) learns to retrieve from \(\mathbf{U}\) the information needed to reconstruct the \(t\)-th acoustic frame. This design turns Mel reconstruction into a sequence-to-sequence generation problem from token-aligned text-prosody representations to frame-aligned acoustic features.

Finally, let \(\mathbf{D}=(\mathbf{d}_1,\dots,\mathbf{d}_L)\), where \(\mathbf{d}_t\in\mathbb{R}^{d_r}\) denotes the decoder output corresponding to the \(t\)-th Mel frame query. Each decoder output vector is then mapped to a Mel frame through a multilayer projection head:
\begin{equation}
\hat{\mathbf{m}}_t = g(\mathbf{d}_t)\in\mathbb{R}^{F},
\label{eq:mel-frame}
\end{equation}
where \(g(\cdot)\) is a feed-forward network and \(F=128\). Collecting all reconstructed frames gives the predicted Mel spectrogram
\begin{equation}
\hat{\mathbf{M}}=(\hat{\mathbf{m}}_1,\dots,\hat{\mathbf{m}}_{L})\in\mathbb{R}^{F\times L}.
\label{eq:mel-spectrogram}
\end{equation}

Overall, the full computation pipeline can be summarized as
\begin{equation}
\mathbf{x}
\rightarrow \mathbf{M}
\rightarrow \mathbf{H}^{\mathrm{enc}}
\rightarrow \{\mathbf{y},\mathbf{P}\}
\rightarrow \mathbf{Z}
\rightarrow \mathbf{U}
\rightarrow \mathbf{D}
\rightarrow \hat{\mathbf{M}}.
\label{eq:full-pipeline}
\end{equation}
This formulation explicitly decomposes spoken input into a semantic stream and a prosodic stream, and then requires the model to jointly use both streams to reconstruct the original acoustics. 

\section{Implementation Details of the Probing Experiments}
\label{apx:probing}
\begin{wraptable}{r}{0.52\textwidth}
\vspace{-1.2\baselineskip}
\centering
\small
\caption{Number of samples in each label of our accent benchmark.}
\label{tab:accent-benchmark-stats}
\vspace{0.5\baselineskip}
\begin{tabular}{@{}p{0.37\textwidth}r@{}}
\toprule
Target label & Samples \\
\midrule
United States English & 310 \\
England English & 310 \\
India and South Asia\newline (India, Pakistan, Sri Lanka) & 309 \\
Europe & 71 \\
\midrule
Total & 1{,}000 \\
\bottomrule
\end{tabular}
\end{wraptable}
For all probing experiments, we train lightweight MLP classifiers on top of the extracted prosody embeddings. We evaluate both a 1-layer and a 2-layer probe. The 1-layer probe consists of a dropout layer followed by a linear classifier, while the 2-layer probe consists of a linear layer, a ReLU activation, a dropout layer, and a final linear classifier. For the 2-layer probe, the hidden dimension is set to 256. Across all probing experiments, we use the Adam optimizer with learning rate $10^{-3}$, dropout rate $0.1$, batch size $32$, and cross-entropy loss. The maximum number of training epochs is set to $20$. We apply early stopping based on validation accuracy with a patience of $3$ epochs, and the checkpoint with the best validation accuracy is used for the final evaluation.

For the evaluation protocol, we do not rely on a single train-test split. Instead, we adopt a stratified multi-seed cross-validation setup. Specifically, for each probe architecture, we run $10$ random seeds and, under each seed, perform $5$-fold stratified cross-validation, resulting in $50$ runs in total per probe. In each run, the data split is determined by shuffled stratified folds, and the model is re-initialized independently. We then aggregate the results across all runs. Therefore, the probing numbers reported in the paper are robust estimates obtained from repeated stratified cross-validation.

\section{Accent Benchmark Construction Details}
\label{apx:accent_benchmark}
We construct the accent benchmark from the English split test set of Common Voice 22.0 by selecting four target labels that are both representative of major English accent groups and sufficiently well supported in Common Voice \cite{commonvoice}: ``United States English'', ``England English'', ``India and South Asia (India, Pakistan, Sri Lanka)'', and ``Europe''. Following AIR-Bench \cite{airbench}, the final benchmark contains 1{,}000 samples, with the per-category distribution summarized in Table~\ref{tab:accent-benchmark-stats}. Because Common Voice does not provide a single label named ``Europe'', we map multiple Europe-related accent strings to this category. All the mapped strings are summarized in Table~\ref{tab:accent-europe-mapping}.

\begin{table}[t]
\centering
\small
\caption{Common Voice accent strings grouped into the ``Europe'' category in our accent benchmark.}
\label{tab:accent-europe-mapping}
\begin{tabular}{p{0.92\linewidth}}
\toprule
Common Voice labels mapped to ``Europe'' \\
\midrule
German; French; Russian; German English; Dutch; Icelandic; Dutch English; Polish; German Accent; French Accent; Slovak; Deutsch English; English with a French accent; German native; native Dutch speaking; Swedish; Finnish \\
\bottomrule
\end{tabular}
\end{table}

\clearpage
\section{\ourencoder{} Design Choice Investigations}
\begin{figure}[htbp]
    \centering
    \includegraphics[width=0.92\textwidth]{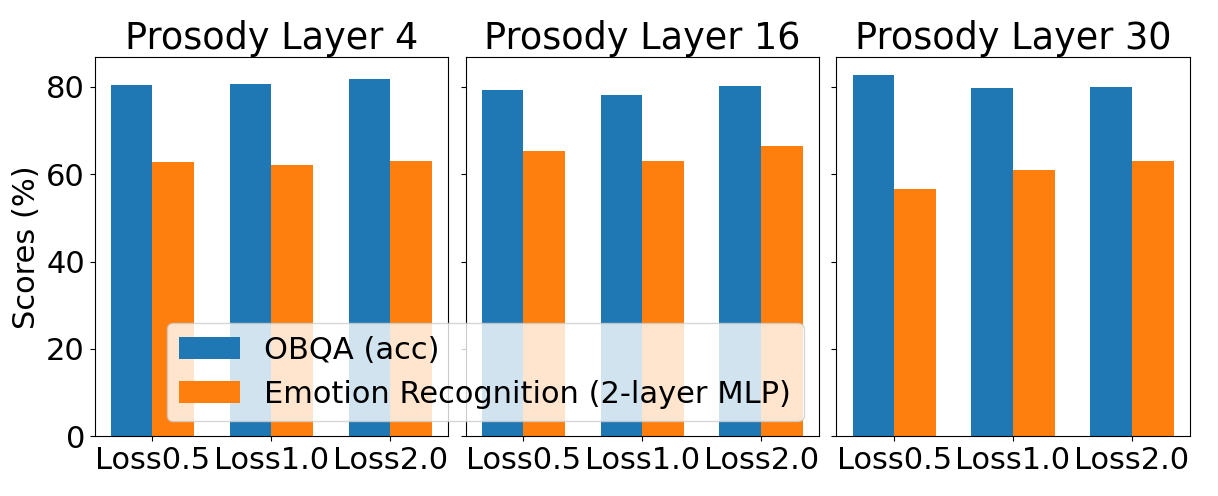}
    \caption{Design choice investigation results for \ourencoder{}. The blue and orange bars represent OpenBookQA accuracy and the 2-layer MLP probing results on the SER task.}
    \label{fig:speech_encoder_ablation}
\end{figure}
We study two central design choices in \ourencoder{}. \textbf{Prosody layer.} We vary which Whisper decoder layer is used to extract the prosody embedding and test representative early, middle, and late layers (layers 4, 16, and 30 out of 32 decoder layers). The reconstruction is trained on the prosody embeddings produced by the corresponding layer. \textbf{Loss ratio ($\lambda$).} We vary the weight that balances the ASR loss and the reconstruction loss, considering $\lambda\in\{0.5, 1.0, 2.0\}$. We evaluate each setting on both a semantic task and a prosodic task. For semantic evaluation, we pair the resulting encoder with Qwen2.5-7B-Instruct and measure question-answering accuracy on OpenBookQA. For prosodic evaluation, we use the 2-layer MLP probing setup on SER. As shown in Figure \ref{fig:speech_encoder_ablation}, performance is broadly stable across these choices, suggesting that \ourencoder{} is robust to moderate changes in both the selected prosody layer and the loss ratio.

\subsection{Why the 16th Decoder Layer?}
\label{apx:decoder-layer-rationale}
We aim to justify the use of the 16th Whisper decoder layer as the source of prosody embeddings. We examine sensitivity to decoder depth and compare encoder-side alternatives to assess whether this choice offers a suitable balance of semantic and paralinguistic information.

The comparison of decoder layers 4, 16, and 30 in Figure~\ref{fig:speech_encoder_ablation} shows broadly stable semantic and prosodic performance. We therefore choose layer 16 as a balanced middle-layer default, rather than because it is uniquely optimal. Our motivation is that early decoder layers may be less contextualized and more sensitive to local acoustic variation, whereas late layers are more directly optimized for ASR next-token prediction. An intermediate layer offers a natural compromise between integrating acoustic information through cross-attention and preceding textual context, and specialization toward lexical prediction. This is a motivation for the design; the empirical results suggest that the method is not sensitive to the exact layer choice.

\paragraph{Encoder-side alternatives.}
We further compare the layer-16 decoder representation with two encoder-side alternatives. \textbf{Encoder Global} applies masked mean pooling to the final encoder hidden states over valid audio frames. \textbf{Encoder-to-Token Aligned} averages the layer-16 decoder cross-attention weights over attention heads and uses these weights to aggregate the final encoder states for each generated token. Table~\ref{tab:encoder-decoder-representations} reports semantic average accuracy and the four AIR-Bench paralinguistic results for the 7B model.

\begin{table}[htbp]
\centering
\caption{Encoder- and decoder-side representations in \ourencoder{}. Semantic Avg. is the average accuracy on the six semantic benchmarks; the remaining columns report AIR-Bench paralinguistic accuracy. All scores are percentages ($\uparrow$).}
\label{tab:encoder-decoder-representations}
\footnotesize
\setlength{\tabcolsep}{3pt}
\begin{tabular}{@{}lrrrrrr@{}}
\toprule
Representation & Semantic Avg. & Emotion & Gender & Age & Accent & Paralinguistic Avg. \\
\midrule
Decoder Layer 16 & 82.50 & 60.50 & 88.60 & 64.90 & 45.10 & 64.80 \\
Encoder Global & 81.83 & 59.30 & 93.22 & 68.70 & 42.80 & 66.01 \\
Encoder-to-Token Aligned & 82.27 & 53.75 & 93.62 & 67.30 & 41.30 & 63.99 \\
\bottomrule
\end{tabular}
\end{table}

The semantic averages are comparable (81.83\% and 82.27\% vs. 82.50\% for the decoder representation). On AIR-Bench, global encoder pooling performs slightly better on average (66.01\% vs. 64.80\%), whereas token-aligned encoder features perform slightly worse (63.99\%). Encoder-side features thus do not provide a consistent downstream advantage, and the additional alignment procedure does not improve performance in this comparison.

\paragraph{Reconstruction-trained, token-synchronized states.}
The selected decoder states are not merely off-the-shelf Whisper features optimized for text prediction. In \ourencoder{}, the Mel reconstruction loss is backpropagated directly through the layer-16 states, encouraging them to retain acoustic and paralinguistic information beyond lexical content. Given comparable downstream performance, natural synchronization with generated text tokens, and no need for an additional alignment procedure, layer-16 decoder states provide a simple and effective default.

\section{LLM Design Choice Investigations}
We investigate the effect of two critical design choices in LLM training.

\textbf{1) Knowledge Distillation Loss Type:} In prior studies that address the modality gap problem, they typically use Kullback-Leibler (KL)-divergence loss for knowledge distillation \cite{salad,cross_modal_distillation}. Especially, SALAD \cite{salad} finds that KL loss achieves significantly better modality alignment results. Therefore, we investigate whether KL loss or cross entropy loss provides better results. Specifically, we use a token-level forward KL objective in which the teacher and student are conditioned on different input sequences:
\begin{equation}
\mathcal{L}_{\mathrm{KL}}
=
T^2 \sum_{j \in \mathcal{A}}
\mathrm{KL}\!\left(
P_t\!\left(\cdot \mid \mathbf{x}^{(t)}_{\le j}\right)
\;\middle\|\;
P_s\!\left(\cdot \mid \mathbf{x}^{(s)}_{\le j}\right)
\right),
\end{equation}
where $\mathcal{A}$ denotes the set of answer-token positions used for distillation, $\mathbf{x}^{(t)}$ and $\mathbf{x}^{(s)}$ denote the teacher and student input sequences, respectively, and $T$ is the distillation temperature. In our setting, the teacher input $\mathbf{x}^{(t)}$ is the clean text-only sequence, while the student input $\mathbf{x}^{(s)}$ is the corresponding sequence containing ASR text tokens and its injected prosody embedding. The predictive distributions are obtained from the teacher and student logits as
\begin{equation}
P_t\!\left(\cdot \mid \mathbf{x}^{(t)}_{\le j}\right)
=
\mathrm{softmax}\!\left(\frac{\mathbf{z}^{(t)}_j}{T}\right),
\qquad
P_s\!\left(\cdot \mid \mathbf{x}^{(s)}_{\le j}\right)
=
\mathrm{softmax}\!\left(\frac{\mathbf{z}^{(s)}_j}{T}\right),
\end{equation}
where $\mathbf{z}^{(t)}_j$ and $\mathbf{z}^{(s)}_j$ are the teacher and student logits at position $j$. This formulation makes explicit that we distill the teacher's next-token distribution into the student's next-token distribution under different conditioning contexts.

In addition to the KL term, our training objective also includes the standard next-token cross-entropy loss. We balance the two losses using a coefficient $\alpha \in [0,1]$, yielding the overall objective
\begin{equation}
\mathcal{L}
=
\alpha \mathcal{L}_{\mathrm{KL}} + (1-\alpha)\mathcal{L}_{\mathrm{CE}},
\end{equation}
where $\mathcal{L}_{\mathrm{CE}}$ denotes the cross-entropy loss on the target tokens and $\mathcal{L}_{\mathrm{KL}}$ is the distillation loss defined above. To study the effect of this trade-off, we evaluate $\alpha \in \{0, 0.5, 1\}$. Here, $\alpha=0$ corresponds to pure cross-entropy training, $\alpha=1$ corresponds to pure KL distillation, and $\alpha=0.5$ gives equal weight to the two objectives. For all distillation experiments, we fix the temperature to $T=2.0$.

\textbf{2) Knowledge Distillation Text Token Source:} Because \ourencoder{} internally performs ASR to produce text tokens, an important design question is which text source should be used during distillation. More specifically, we ask whether the student should be trained with ground-truth text, with ASR-transcribed text, or with a mixture in which the student input and teacher target are derived from different text sources. To answer this question, we compare the following three settings:
\begin{enumerate}[left=0pt]
  \item \textbf{GTQ\_GTA:} The student input uses the original ground-truth text, and the teacher distribution is also computed from the original ground-truth text. This is the cleanest setting, and it corresponds to distillation under an idealized assumption of error-free transcription.
  \item \textbf{ASRQ\_ASRA:} The student input uses the text transcribed by \ourencoder{} from the speech-form data, and the teacher distribution is also conditioned on this transcribed text. This setting matches inference-time conditions most closely, because the student is trained on the same type of text tokens it will receive at test time. However, the distillation target is also based on text that may contain ASR errors.
  \item \textbf{ASRQ\_GTA:} The student input uses the text transcribed by \ourencoder{}, while the teacher distribution is computed from the original ground-truth text. This setting exposes the student to realistic ASR-imperfect inputs, but still provides a clean teacher target. Intuitively, it tests whether the model can learn to recover from typical ASR errors by mapping noisy transcribed inputs toward the behavior induced by the correct text.
\end{enumerate}
For all three settings, the student always receives the same prosody embeddings in addition to the text tokens during training. Additionally, we use only the knowledge distillation datasets (CommonsenseQA and UltraChat) for training. We evaluate all settings on MMSU and OBQA benchmarks. Regardless of the text source used during training, all text tokens at inference time are produced by \ourencoder{}. As shown in Table~\ref{tab:kd-text-source-alpha}, we do not observe large performance differences across the three settings. However, the ASRQ\_ASRA setting with $\alpha=0$ achieves the best overall results across the two benchmarks, so we adopt this configuration in the final model training.

\begin{table}[t]
\centering
\footnotesize
\caption{Effect of loss weight $\alpha$ and text-token source on MMSU and OBQA (\%).}
\label{tab:kd-text-source-alpha}
{\setlength{\tabcolsep}{4pt}%
\begin{tabular}{@{}llcc@{}}
\toprule
Setting & $\alpha$ & MMSU & OBQA \\
\midrule
GTQ\_GTA & 0.0 & 67.1 & 82.9 \\
GTQ\_GTA & 0.5 & 63.0 & 80.9 \\
GTQ\_GTA & 1.0 & 66.5 & 82.2 \\
\midrule
ASRQ\_ASRA & 0.0 & 68.0 & 84.4 \\
ASRQ\_ASRA & 0.5 & 66.5 & 81.5 \\
ASRQ\_ASRA & 1.0 & 63.0 & 79.8 \\
\midrule
ASRQ\_GTA & 0.0 & 67.6 & 82.4 \\
ASRQ\_GTA & 0.5 & 65.0 & 81.1 \\
ASRQ\_GTA & 1.0 & 63.7 & 81.1 \\
\bottomrule
\end{tabular}%
}
\end{table}

\section{Roles of the Transcript, Semantic Distillation, and Prosody Embeddings}
\label{apx:component-functions}
We aim to clarify which components preserve semantic capabilities and which enable paralinguistic understanding. In particular, we test whether the transcript-compatible interface and semantic distillation account for the small semantic modality gap, while prosody embeddings provide complementary non-lexical information.

Tables~\ref{tab:component-semantic} and~\ref{tab:component-paralinguistic} isolate the roles of the transcript-compatible interface, semantic knowledge distillation, and prosody embeddings through controlled semantic and paralinguistic evaluations.

\paragraph{Controlled settings.}
\textbf{Original Qwen} is the unmodified Qwen2.5-7B-Instruct model evaluated on text inputs. \textbf{Text Semantic Distillation} applies the semantic distillation to text-only Qwen2.5-7B-Instruct using clean text--response data; this control is also referred to as \emph{Text SFT}. \textbf{No Prosody Embedding} uses the same data and training protocol as \ourmethod{}, but omits the prosody stream and trains on \ourencoder{} ASR transcripts with matched ASR-conditioned semantic targets. \textbf{Original Whisper} replaces \ourencoder{} with the original Whisper-large-v3 and trains Qwen using that model's ASR transcripts and layer-16 decoder embeddings, with the same data mixture and training recipe; this is the w/o Recon setting in Table~\ref{tab:paralinguistic}. Finally, \textbf{w/o Semantic Distillation} removes semantic distillation from \ourmethod{} training while retaining paralinguistic-task training.

\begin{table}[htbp]
\centering
\caption{Controlled component ablations: semantic accuracy $\uparrow$ (\%) and gap $\downarrow$ (\%). Each gap is Original Qwen accuracy minus the row's accuracy; negative values indicate improvement over the text baseline. Original Qwen scores are 89.2\%, 70.9\%, 85.8\%, 96.1\%, 89.2\%, and 68.1\%, respectively (Avg.: 83.2\%). Gaps are computed from the reported rounded scores, including the reported averages. For the text-only distillation control, Gap denotes a text-baseline difference, not a speech--text modality gap.}
\label{tab:component-semantic}
\fontsize{8.5}{10}\selectfont
\setlength{\tabcolsep}{0.8pt}
\begin{tabular}{@{}l*{14}{r}@{}}
\toprule
& \multicolumn{2}{c}{StoryCloze} & \multicolumn{2}{c}{MMSU} & \multicolumn{2}{c}{OBQA} & \multicolumn{2}{c}{ARC-E} & \multicolumn{2}{c}{ARC-C} & \multicolumn{2}{c}{PIQA} & \multicolumn{2}{c}{Avg.} \\
\cmidrule(lr){2-3}\cmidrule(lr){4-5}\cmidrule(lr){6-7}\cmidrule(lr){8-9}\cmidrule(lr){10-11}\cmidrule(lr){12-13}\cmidrule(l){14-15}
Model & Acc. & Gap & Acc. & Gap & Acc. & Gap & Acc. & Gap & Acc. & Gap & Acc. & Gap & Acc. & Gap \\
\midrule
Text Semantic Distillation & 93.3 & \GapSC{-4.1} & 71.6 & \GapMMSU{-0.7} & 85.7 & \GapOBQA{0.1} & 96.8 & \GapARCE{-0.7} & 91.6 & \GapARCC{-2.4} & 70.9 & \GapPIQA{-2.8} & 85.0 & \GapAVG{-1.8} \\
No Prosody Embedding & 89.6 & \GapSC{-0.4} & 68.5 & \GapMMSU{2.4} & 82.4 & \GapOBQA{3.4} & 95.7 & \GapARCE{0.4} & 89.4 & \GapARCC{-0.2} & 66.0 & \GapPIQA{2.1} & 81.9 & \GapAVG{1.3} \\
Original Whisper & 81.9 & \GapSC{7.3} & 67.6 & \GapMMSU{3.3} & 83.1 & \GapOBQA{2.7} & 95.8 & \GapARCE{0.3} & 88.8 & \GapARCC{0.4} & 76.9 & \GapPIQA{-8.8} & 82.3 & \GapAVG{0.9} \\
\ourmethod{}-7B w/o Semantic Distillation & 84.7 & \GapSC{4.5} & 46.7 & \GapMMSU{24.2} & 55.8 & \GapOBQA{30.0} & 89.5 & \GapARCE{6.6} & 75.8 & \GapARCC{13.4} & 66.6 & \GapPIQA{1.5} & 69.9 & \GapAVG{13.3} \\
\ourmethod{}-7B & 88.6 & \GapSC{0.6} & 66.7 & \GapMMSU{4.2} & 85.1 & \GapOBQA{0.7} & 95.8 & \GapARCE{0.3} & 89.4 & \GapARCC{-0.2} & 69.7 & \GapPIQA{-1.6} & 82.5 & \GapAVG{0.7} \\
\bottomrule
\end{tabular}
\end{table}

\begin{table}[htbp]
\centering
\caption{Controlled component ablations: AIR-Bench paralinguistic accuracy (\%, $\uparrow$) and the reported four-task average.}
\label{tab:component-paralinguistic}
\normalsize
\setlength{\tabcolsep}{9pt}
\renewcommand{\arraystretch}{1.15}
\begin{tabular}{@{}lrrrrr@{}}
\toprule
Model & Emotion & Gender & Age & Accent & Avg. \\
\midrule
No Prosody Embedding & 54.10 & 38.30 & 29.60 & 30.50 & 38.10 \\
Original Whisper & 47.25 & 77.66 & 55.50 & 32.20 & 53.15 \\
\ourmethod{}-7B & 60.50 & 88.60 & 64.90 & 45.10 & 64.80 \\
\bottomrule
\end{tabular}
\end{table}

\paragraph{The transcript carries most semantic information.}
Without prosody embeddings, semantic accuracy remains close to the full model (81.9\% vs. 82.5\%). Replacing \ourencoder{} with Original Whisper likewise leaves semantic accuracy broadly comparable (82.3\% vs. 82.5\%). These results suggest that the transcript-compatible interface, rather than the prosody embeddings, is the primary source of content-based QA performance on these benchmarks.

\paragraph{Semantic distillation preserves and adapts pretrained capabilities.}
The text-only distillation control achieves 85.0\%, compared with 83.2\% for Original Qwen. By contrast, removing semantic distillation from \ourmethod{} reduces semantic accuracy from 82.5\% to 69.9\%. This contrast supports the role of distillation in preventing degradation and adapting the pretrained TLM during speech and paralinguistic training, rather than attributing the reduced modality gap primarily to a large increase in the underlying text model's capability.

\paragraph{Prosody embeddings supply complementary paralinguistic information.}
Evaluating the same No Prosody Embedding model on AIR-Bench reduces average accuracy from 64.8\% to 38.1\%, a 26.7-point drop despite the small semantic difference. Original Whisper reaches 53.15\% vs. 64.80\% for \ourmethod{}, an 11.65-point drop, while maintaining similar semantic accuracy. Thus, transcripts and distillation support semantics, while reconstruction-trained prosody embeddings supply complementary paralinguistic information.

\section{More Related Works}
Several prior works also factorize or decouple different aspects of speech, but they are largely motivated by 
\emph{generation quality}, \emph{prosody-rich synthesis}, or \emph{prosody control}, rather than by reducing the semantic modality gap between a TLM and an SLM. For example, AudioLM \cite{audiolm}, pGSLM \cite{pgslm}, SpiritLM \cite{spiritlm}, Moshi \cite{moshi}, and SpeechGPT-Gen \cite{speechgpt-gen} separate semantic/content information from lower-level acoustic, pitch, style, or codec streams so that the model can better generate natural and expressive speech. A related line of work focuses more explicitly on controllable generation. NaturalSpeech~3 \cite{naturalspeech3} factorizes speech into multiple codec subspaces such as content and prosody and then generates them with a cascade of diffusion modules, while ProsodyLM \cite{prosodylm} converts each utterance into text together with discrete word-level prosody tokens so that an LLM can better model prosodic phenomena. These approaches show the value of disentangling content from other speech attributes, but their main objective is still to improve spoken generation or prosody control.

Our motivation is different. We are not primarily trying to make the decoder produce more expressive speech; instead, we seek to make spoken \emph{input} look more like the input of a text LLM while preserving paralinguistic information needed for understanding. Concretely, \ourmethod{} uses synchronized text tokens and compact continuous prosody embeddings, so the LLM can retain the semantic strengths of the original TLM while gaining stronger paralinguistic understanding. This differs from prior token-factorization approaches that represent prosody or acoustics as additional discrete streams or interleaved tokens, which are well suited for generation but can substantially lengthen the sequence and shift the interaction format away from standard TLM processing. In this sense, our method is closer to an input-side alignment strategy: the goal is better semantic preservation under speech input, rather than richer speech synthesis alone.

\paragraph{Comparison with DrVoice.}
DrVoice \cite{drvoice} is closely related to our input-side perspective: it groups 25\,Hz speech representations into a 5\,Hz sequence, reducing the temporal-resolution mismatch between speech and text. In contrast, \ourmethod{} explicitly decomposes speech into transcript tokens and compact prosody embeddings, addressing both sequence-length and representation-format mismatches with the native text inputs of the underlying TLM.

On the shared VoiceBench tasks, DrVoice reports 82.20\% on OpenBookQA and 67.27\% on MMSU, compared with 85.1\% and 66.7\%, respectively, for \ourmethod{}-7B, indicating broadly comparable performance on these two tasks. These are comparisons of reported results rather than a controlled evaluation under matched training and inference conditions. DrVoice uses approximately 56K hours of speech in post-training and additionally applies DPO and GSPO to improve speech comprehension, instruction following, and reasoning \cite{drvoice}. Such additional optimization can change the capabilities of the underlying text checkpoint, complicating an interpretation solely in terms of modality-gap reduction. By comparison, \ourmethod{} uses roughly 1K hours of LLM training audio with supervised semantic distillation and paralinguistic training, without additional preference optimization or reinforcement learning. These data figures concern model adaptation rather than the total data used to pretrain the underlying components. The results suggest that text--prosody factorization can support competitive speech understanding with limited LLM adaptation data, while the different training pipelines preclude attributing the observed differences solely to the input representation.

\section{Broader Societal Impact}
\label{apx:broader_impact}
Our work studies how to reduce the modality gap in SLMs by improving the input-side speech representation rather than relying only on output-side adaptation. A potential positive societal impact of this direction is that it may make speech-based AI systems substantially more semantically capable, allowing users to access stronger reasoning, knowledge use, and instruction-following abilities directly through spoken interaction. In settings such as voice assistants, spoken question answering, accessibility tools, and hands-free interfaces, reducing the modality gap could make speech systems more reliable for general tasks, especially when current models lose important meaning when moving from text input to speech input.

At the same time, making SLMs more semantically powerful can introduce meaningful risks. First, if a model becomes better at extracting task-relevant meaning from speech, users may place greater trust in spoken interfaces even when they still make subtle but consequential mistakes in factual understanding, reasoning, or instruction interpretation; such failures could be especially harmful in high-stakes domains. Second, more capable speech interfaces may accelerate deployment of automated systems in education, customer service, or other communication-heavy settings, potentially amplifying harms from misinformation, over-reliance, or reduced human oversight on a larger scale. Third, although not the primary focus of our work, better use of speech-side cues such as prosody may also increase the possibility that systems infer sensitive speaker attributes or intentions in ways users do not expect, raising privacy and fairness concerns if such capabilities are deployed carelessly.

\begin{table}[t]
\centering
\small
\caption{Summary of dataset licenses and whether academic research use is allowed.}
\label{tab:dataset-license-summary}
\begin{tabular}{@{}lll@{}}
\toprule
Dataset & License & Academic research use \\
\midrule
CommonsenseQA & MIT & Yes \\
LibriSpeech & CC BY 4.0 & Yes \\
UltraChat & MIT & Yes \\
ParaSpeechCaps & CC BY-NC-SA 4.0 & Yes \\
IEMOCAP & Custom research-only license & Yes \\
CREMA-D & ODbL 1.0 / DbCL 1.0 & Yes \\
SAVEE & Custom institutional license & Yes \\
TESS & CC BY-NC-ND 4.0 & Yes \\
ESD & Custom non-commercial license & Yes \\
Common Voice & CC0 & Yes \\
\bottomrule
\end{tabular}
\end{table}


\end{document}